\documentclass{article}

\usepackage[final]{nips_2018}

\usepackage[utf8]{inputenc} 
\usepackage[T1]{fontenc}    
\usepackage{hyperref}       
\usepackage{url}            
\usepackage{booktabs}       
\usepackage{amsfonts}       
\usepackage{nicefrac}       
\usepackage{microtype}      

\usepackage{wrapfig}
\usepackage{graphicx}

\newtheorem{definition}{Definition}

\usepackage[colorinlistoftodos]{todonotes}

\usepackage{float}
\usepackage{multirow}

\usepackage{amsmath}
\usepackage{mleftright}

\usepackage{bm}
\usepackage{bbm}

\usepackage{threeparttable}

\usepackage{subfig}

\usepackage{xspace}
\newcommand{\model}{GSimCNN\xspace}

\begin{document}

\newcommand*\samethanks[1][\value{footnote}]{\footnotemark[#1]}

\title{Convolutional Set Matching for Graph Similarity}
\author{
  Yunsheng Bai\thanks{The two first authors made equal contributions.}\space,\textsuperscript{1}  Hao Ding\samethanks \space,\textsuperscript{2} Yizhou Sun,\textsuperscript{1} Wei Wang\textsuperscript{1} \\
  \textsuperscript{1}University of California, Los Angeles, \textsuperscript{2}Purdue University \\
  \texttt{yba@ucla.edu, ding209@purdue.edu, yzsun@cs.ucla.edu, weiwang@cs.ucla.edu} \\
}
\maketitle
\section{A Multi-Scale Convolutional Model for Pairwise Graph Similarity}

\begin{wrapfigure}{r}{0.5\textwidth}
\vspace{-1em}
\centering
  \includegraphics[width=0.435\textwidth]{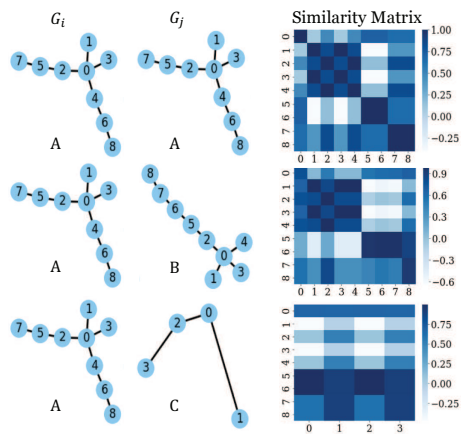}
\caption{Illustration of three graph-graph similarity matrices generated by our end-to-end \model trained on the LINUX dataset~\cite{wang2012efficient}. Nodes are ordered and labelled with their ids, and darker colors indicate greater similarities between nodes. Convolutional Neural Networks are applied to these matrices to generate the graph-graph similarity score.}
\label{fig:vgae}
\vspace{-1.8em}
\end{wrapfigure}

We introduce GSimCNN \emph{\textbf{}} (\emph{\underline{G}}raph \emph{\underline{Sim}}ilarity Computation via \emph{\underline{C}}onvolutional \emph{\underline{N}}eural \emph{\underline{N}}etworks) for predicting the similarity score between two graphs. As the core operation of graph similarity search, pairwise graph similarity computation is a challenging problem due to the NP-hard nature of computing many graph distance/similarity metrics. 

We demonstrate our model using the Graph Edit Distance (GED)~\cite{bunke1983distance} as the example metric. It is defined as the number of edit operations in the optimal alignments that transform one graph into the other, where an edit operation can be an insertion or a deletion of a node/edge, or relabelling of a node. It is NP-hard~\cite{zeng2009comparing} and costly to compute in practice~\cite{blumenthal2018exact}.

The key idea is to turn the pairwise graph distance computation problem into a learning problem. This new approach not only offers parallelizability and efficiency due to the nature of neural computation, but also achieves significant improvement over state-of-the-art GED approximation algorithms. 

\textbf{Definitions} \enspace We are given an undirected, unweighted graph $\mathcal{G}=(\mathcal{V},\mathcal{E})$ with $N=|\mathcal{V}|$ nodes. Node features are summarized in an $N\times D$ matrix $\bm{H}$. We transform GED into a similarity metric ranging between 0 and 1. Our goal is to learn a neural network based function that takes two graphs as input and outputs the similarity score that can be transformed back to GED through a one-to-one mapping.

\textbf{\model} \enspace \model consists of the following sequential stages: 1) \textit{Multi-Scale Graph Convolutional Network~\cite{kipf2016semi} layers} generate vector representations for each node in the two graphs at different scales; 2) \textit{Graph Interaction layers} compute the inner products between the embeddings of every pair of nodes in the two graphs, resulting in multiple similarity matrices capturing the node-node interaction scores at different scales; 3) \textit{Convolutional Neural Network layers} convert the similarity computation problem into a pattern recognition problem, which provides multi-scale features to a \textit{fully connected network} to obtain a final predicted graph-graph similarity score. An overview of our model is illustrated in Fig.~\ref{fig:model}. 

\subsection{Stage I:  Multi-Scale GCN Layers}

\begin{figure}
\vspace{-1em}
\centering
  \includegraphics[width=1.0\textwidth]{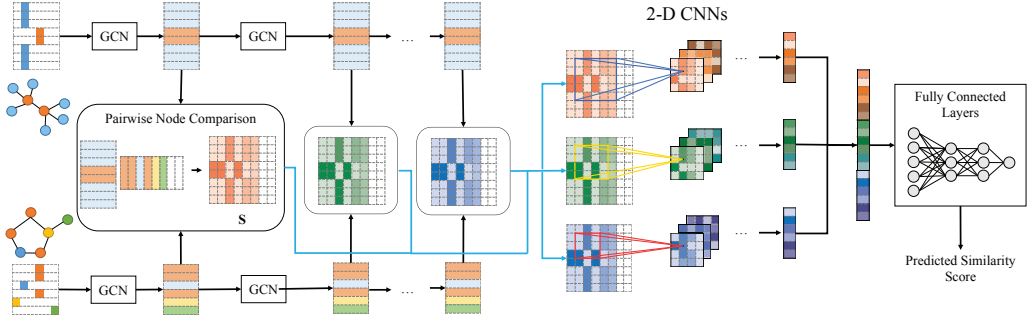}
\caption{An overview illustration. Three similarity matrices are generated with three 2-D CNNs.}
\label{fig:model}
\vspace{-1.5em}
\end{figure}

In Stage I, we generate node embeddings by multi-layer GCNs, where each layer is defined as~\cite{kipf2016semi}: 
\begin{equation} \label{eq:gcn} \mathrm{conv}(\bm{x_i}) = \mathrm{ReLU}( \sum_{j \in \mathcal{N}(i)} \frac{1}{\sqrt[]{d_i d_j}} \bm{x}_j \bm{W}^{(l)} + \bm{b}^{(l)})\end{equation}
Here, $\mathcal{N}(i)$ is the set of all first-order neighbors of node $i$ plus node $i$ itself; $d_i$ is the degree of node $i$ plus 1; $\bm{W}^{(l)} \in \mathbb{R}^{D^{l} \times D^{l+1}}$ is the weight matrix associated with the $n$-th GCN layer; $\bm{b}^{(l)} \in \mathbb{R}^{D^{l+1}}$ is the bias; and $\mathrm{ReLU}(x)=\mathrm{max}(0,x)$ is the activation function.

In Fig.~\ref{fig:model}, different node types are represented by different colors and one-hot encoded as the initial node representation. For graphs with unlabeled nodes, we use the same constant vector as the initial representation. As shown in \cite{kipf2016variational} and \cite{hamilton2017inductive}, the graph convolution operation aggregates the features from the first-order neighbors of the node. Stacking $N$ GCN layers would enable the final representation of a node to include its $N$-th order neighbors. 

\textbf{Multi-Scale GCN} \enspace The potential issue of using a deep GCN structure is that the embeddings may be too coarse after aggregating neighbors from multiple scales. The problem is especially severe when the two graphs are very similar, as the differences mainly lie in small substructures. Due to the fact that structural difference may occur at different scales, we extract the output of each GCN layer and construct multi-scale interaction matrices, which will be described in the next stage.

\subsection{Stage II: Graph Interaction Layers}

We calculate the inner products between all possible pairs of node embeddings between two graphs from different GCN layers, resulting in multiple similarity matrices $\bm{S}$. Since we later use CNNs to process these matrices, we utilize the breadth-first-search (BFS) node-ordering scheme proposed in \cite{you2018graphrnn} to reorder the node embeddings, running in quadratic time in the worst case. 

\textbf{Max Padding} \enspace Suppose $\mathcal{G}_1$ and $\mathcal{G}_2$ contain $N_1$ and $N_2$ nodes, respectively. To reflect the difference in graph sizes in the similarity matrix, we pad $|N_1 - N_2|$ rows of zeros to the node embedding matrix of the smaller of the two graphs, so that both graphs contain $\mathrm{max}(N_1,N_2)$ nodes. 

\textbf{Matrix Resizing} \enspace
To apply CNNs to the similarity matrices, we apply bilinear interpolation, an image resampling technique~\cite{thevenaz2000image} to resize every similarity matrix. The resulting similarity matrix $\bm{S}$ has fixed shape $M \times M$, where $M$ is a hyperparameter controlling the degree of loss of information due to the resampling.

The following equation summarizes a single Graph Interaction Layer:
\begin{equation} \label{eq:gil} \bm{S} = \mathrm{RES}_{M}(\widetilde{\bm{H}}_1 \widetilde{\bm{H}}_2^{T})\end{equation}
where $\widetilde{\bm{H}_i} \in \mathbb{R}^{\mathrm{max}(N_1,N_2) \times D}, i \in \{1,2\}$ is the padded node embedding matrix $\bm{H}_i \in \mathbb{R}^{N_i \times D}, i \in \{1,2\}$ with zero or $|N_1 - N_2|$ nodes padded, and $\mathrm{RES}(\cdot): \mathbb{R}^{\mathrm{max}(N_1,N_2) \times \mathrm{max}(N_1,N_2)} \mapsto \mathbb{R}^{M \times M}$ is the resizing function.

\subsection{Stage III: CNN and Dense Layers}

The similarity matrices at different scales are processed by multiple independent CNNs, turning the task of graph similarity measurement into an image processing problem. The filter of CNN detect the optimal node matching pattern in the image, and max pooling in CNN select the best matching. The CNN results are concatenated and fed into multiple fully connected layers, so that a final similarity score $\hat{s}_{ij}$ is generated for the graph pair $\mathcal{G}_i$ and $\mathcal{G}_j$. The mean squared error loss function is used to train the model.

\section{Set Matching Based Graph Similarity Computation}

Through GCN transformation, \model encodes the link structure around each node into its vector representation, and thus regards a graph as a set of node embeddings. It essentially reduces the link structure, and simplifies the graph similarity/distance computation into matching two sets. In this section, we formally define the general approach of using set matching to compute graph similarity/distance, and provide detailed theoretical analysis in Appendix A.

\begin{definition}{Graph transforming function:}
\label{def:gtf}
   A graph transforming function $f(\cdot)$ transforms a graph $\mathcal{G}$ into a set of objects, $M$.
\end{definition}

\begin{definition}{Set matching function:}
\label{def:gtf}
   A set matching function $g(\cdot,\cdot)$ takes two sets as input, and returns a score denoting the degree of matching between the two input sets.
\end{definition}

In fact, the forward pass of \model can be interpreted as a two-step procedure: 1. Applying a GCN-based graph transforming function; 2. Applying a CNN-based set matching function. The Appendix A furnishes the comparisons with two types of graph distance algorithms, which would shed light on why \model works effectively.

\section{Experiments on Graph Similarity Search}

\begin{wrapfigure}{r}{0.55\textwidth}
\vspace{-1.8em}
    \centering
    \subfloat[On AIDS. Different colors represent different node labels.]
    {{\includegraphics[scale=0.37]{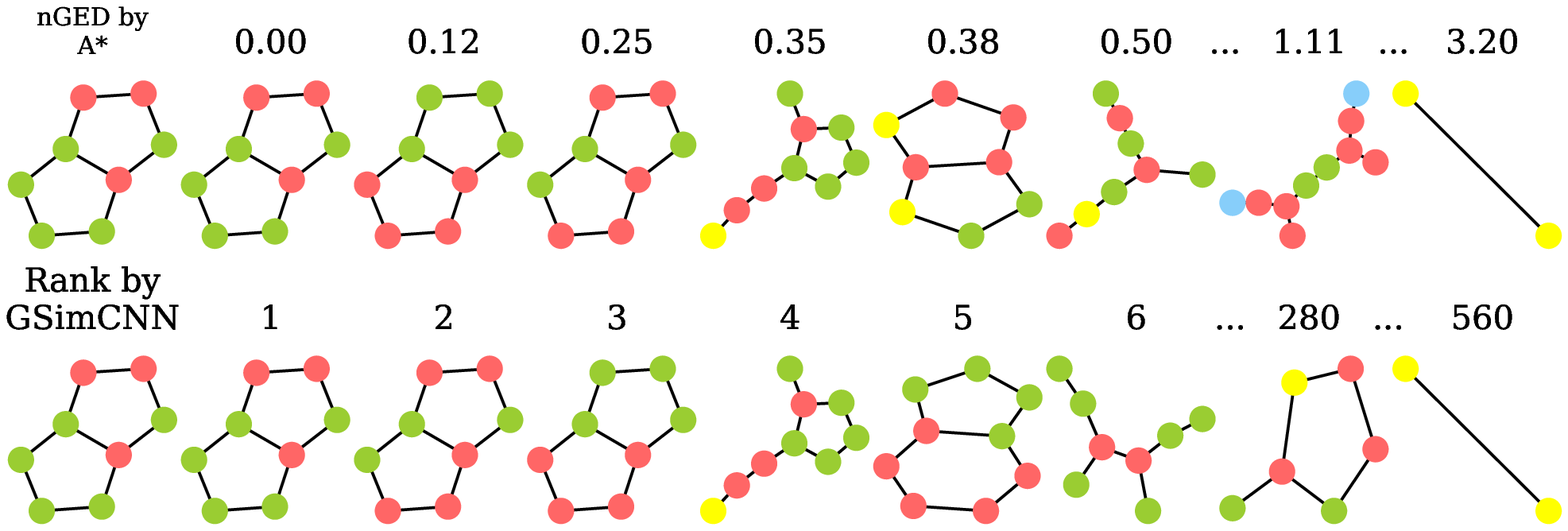} }}
    \vspace{-0.0in}
    \centering
    \subfloat[On LINUX.]
    {{\includegraphics[scale=0.37]{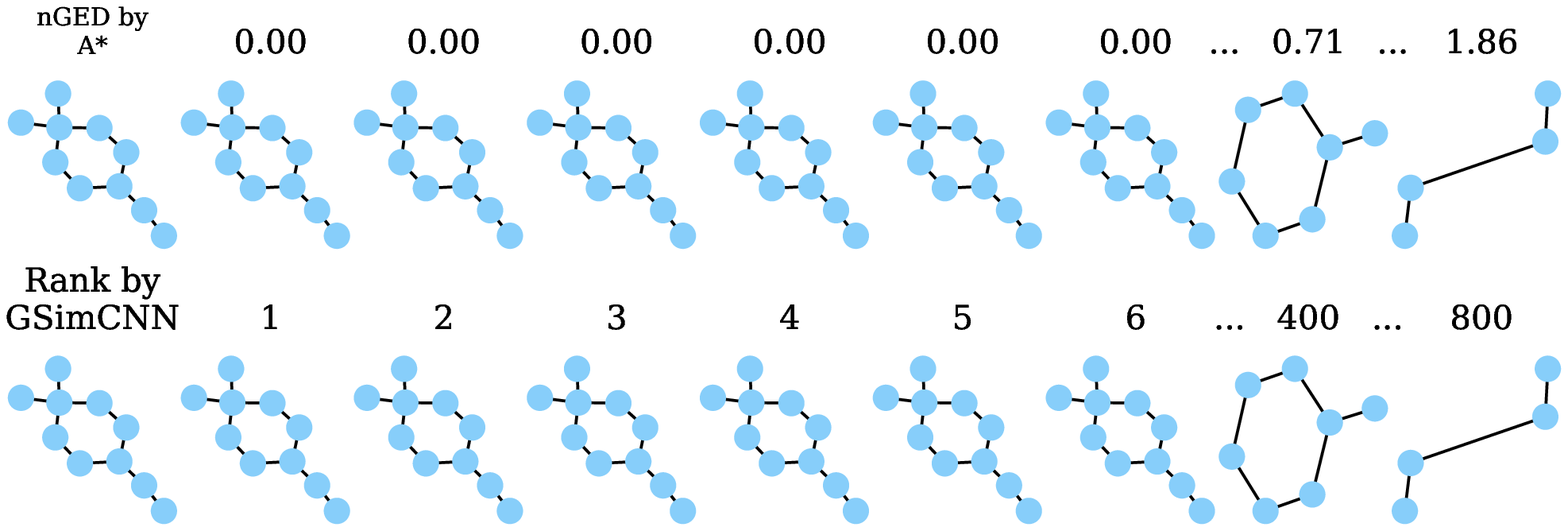} }}
    \vspace{-0.0in}
    \centering
    \subfloat[On IMDB.]
    {{\includegraphics[scale=0.37]{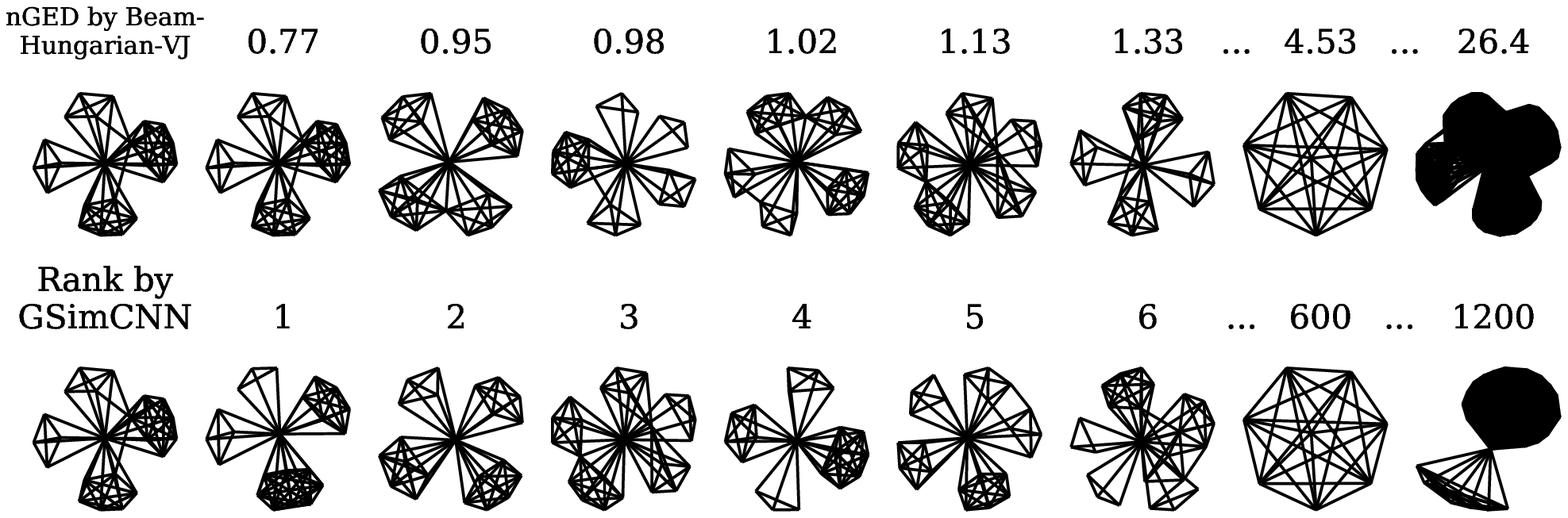} }}
    \vspace{-0.0in}
    \caption{Query case studies.}
    \label{fig:search_demo}
\vspace{-2.8em}    
\end{wrapfigure}

Graph similarity search is among the most important graph-based applications, e.g. finding the chemical compounds that are most similar to a query compound. The goal of these experiments is to demonstrate that \model can alleviate the computational burden while preserving a good performance of GED approximation. We train \model on three real graph datasets~\cite{zeng2009comparing,wang2012efficient,yanardag2015deep}, whose details~\footnote{We make the datasets used in this paper publicly available at \url{https://drive.google.com/drive/folders/1BFj66jqzR_VlWgASEfNMHwAQZ967HV0W?usp=sharing}.} can be found in Appendix B.

We compare methods based on their ability to correctly compute the pairwise graph similarity and rank the database graphs for user query graphs. The training and validation sets contain 60\% and 20\% of graphs, respectively, and serve as the database graphs. The validation set is used for optimization of hyperparameters. The test set contains 20\% of graphs, treated as the query graphs. 

We compare against two sets of baselines: (1) Combinatorial optimization-based algorithms for approximate GED computation: Beam~\cite{neuhaus2006fast}, VJ~\cite{fankhauser2011speeding}, Hungarian~\cite{riesen2009approximate}, HED~\cite{fischer2015approximation}; (2) Neural Network based models: Siamese MPNN~\cite{ribalearning}, EebAvg, GCNMean, GCNMax~\cite{defferrard2016convolutional} (see the Appendix C for details).

To transform ground-truth GEDs into ground-truth similarity scores to train our model, we first normalize the GEDs: $\mathrm{nGED}(\mathcal{G}_1,\mathcal{G}_2)=\frac{\mathrm{GED}(\mathcal{G}_1,\mathcal{G}_2)} {(|\mathcal{G}_1| + |\mathcal{G}_2|) / 2}$, where $|\mathcal{G}_i|$ denotes the number of nodes of $\mathcal{G}_i$~\cite{qureshi2007graph}, and then adopt the exponential function $\lambda(x) = e^{-x}$, an one-to-one function, to transform the normalized GED into a similarity score in the range of $(0, 1]$.

\textbf{Effectiveness} \enspace 
The results on the three datasets can be found in Table~\ref{results}. We report \textit{Mean Squared Error (mse)}, \textit{Kendall's Rank Correlation Coefficient ($\tau$)}~\cite{kendall1938new} and \textit{Precision at $k$ (p@$k$)} for each model on the test set. As shown in Fig.~\ref{fig:search_demo}. In each demo, the top row depicts the query along with the ground-truth ranking results, labeled with their normalized GEDs to the query. The bottom row shows the graphs returned by our model, each with its rank shown at the top. \model is able to retrieve graphs similar to the query.

\begin{table}
\small
\centering
\vspace{-0.05in}
\caption{Results on AIDS, LINUX and IMDB. mse is in $10^{-3}$. The results are based on the split ratio of 6:2:2. We repeated 10 times on AIDS, and the standard deviation of mse is $4.56*10^{-5}$.}
  \begin{tabular}{cccccccccc}
    \toprule
    \multirow{3}[-3]{1.2cm}{\textbf{Method}} &
      \multicolumn{3}{c}{\textbf{AIDS}} &
      \multicolumn{3}{c}{\textbf{LINUX}} &
      \multicolumn{3}{c}{\textbf{IMDB}} \\
      & {mse} & {$\tau$} & {p@10} & {mse} & {$\tau$} & {p@10} & {mse} & {$\tau$} & {p@10}\\
      \midrule
    A* & 0.000* & 1.000* & 1.000* & 0.000* & 1.000* & 1.000* & - & - & - \\
    Beam & 12.090 & 0.463 & 0.481 & 9.268 & 0.714 & 0.973 & 2.413* & 0.837* & 0.803* \\
    Hungarian & 25.296 & 0.378 & 0.360 & 29.805 & 0.517 & 0.913 & 1.845* & 0.872* & 0.825* \\
    VJ & 29.157 & 0.383 & 0.310 & 63.863 & 0.450 & 0.287 & 1.831* & 0.874* & 0.815* \\
    HED & 28.925 & 0.469 & 0.386 & 19.553 & 0.801 & 0.982& 19.400 & 0.627 & 0.801 \\
    \midrule
    Siamese MPNN & 5.184 & 0.210 & 0.032 & 11.737 & 0.024 & 0.009 & 32.596 & 0.093 & 0.023 \\
    EmbAvg & 3.642 & 0.455 & 0.176 & 18.274 & 0.012 & 0.071 & 71.789 & 0.179 & 0.233 \\
    GCNMean & 3.352 & 0.501 & 0.186 & 8.458 & 0.424 & 0.141 & 68.823 & 0.307 & 0.200 \\
    GCNMax & 3.602 & 0.480 & 0.195 & 6.403 & 0.495 & 0.437 & 50.878 & 0.342 & 0.425 \\
    \midrule
    \model & \textbf{0.787} & \textbf{0.724} & \textbf{0.534} & \textbf{0.058} & \textbf{0.962} & \textbf{0.992} & \textbf{0.743} & \textbf{0.847} & \textbf{0.828} \\
    \bottomrule
  \end{tabular}
  \begin{tablenotes}
  \item \footnotesize * On AIDS and LINUX, A* is used as the ground truth. On the largest dataset, IMDB, A* runs too slow; Since Beam, Hungarian, and VJ are guaranteed to return upper bounds to the exact GEDs, we take the minimum of the three as the ground truth. This approach has been adopted by the ICPR 2016 Graph Distance Contest: \url{https://gdc2016.greyc.fr/}.
  \end{tablenotes}
  \vspace{-2.0em}
\centering
\label{results}
\end{table}

\textbf{Efficiency} \enspace In Fig.~\ref{fig:time}, the results are averaged across queries and in wall time. EmbAvg is the fastest method among all, but its performance is poor, since it simply takes the dot product between two graph-level embeddings (average of node embeddings) as the predicted similarity score. Beam and Hungarian run fast on LINUX, but due to their higher time complexity as shown in Table 2, they scale poorly on the largest dataset, IMDB. In general, neural network based models benefit from the parallelizability and acceleration provided by GPU, and in particular, our model \model achieves the best trade-off between running time and performance.

\begin{minipage}{\textwidth}
  \begin{minipage}[b]{0.61\textwidth}
    \centering
    \includegraphics[scale=0.149]{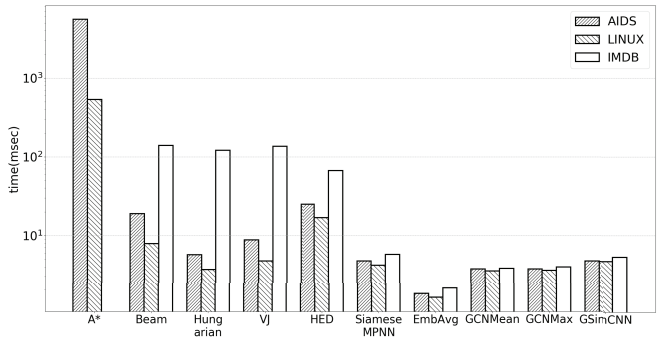}
    \captionof{figure}{Running time comparison.}
    \label{fig:time}
  \end{minipage}
  \hfill
  \begin{minipage}[b]{0.37\textwidth}
    \centering
    \scriptsize
    \begin{tabular}
    {cc} \hline
    \textbf{Method} & \textbf{Time Complexity} \\ \hline
    \textbf{A*}~\cite{hart1968formal} & {$O({N_1}^{N_2})$} \\
    \textbf{Beam}~\cite{neuhaus2006fast} & {subexponential} \\
    \textbf{Hungarian}~\cite{riesen2009approximate} & {$O(({N_1}+{N_2})^3)$} \\ 
    \textbf{VJ}~\cite{fankhauser2011speeding} & {$O(({N_1}+{N_2})^3)$} \\
    \textbf{HED}~\cite{fischer2015approximation} & {$O(({N_1}+{N_2})^2)$} \\
    \textbf{Siamese MPNN}~\cite{ribalearning} & {$O(\mathrm{max}(E_1, E_2, N_1 N_2))$} \\
    \textbf{EmbAvg} & {$O(\mathrm{max}(E_1, E_2))$} \\
    \textbf{GCNMean}~\cite{kipf2016semi} & {$O(\mathrm{max}(E_1, E_2))$} \\
    \textbf{GCNMax}~\cite{kipf2016semi} & {$O(\mathrm{max}(E_1, E_2))$} \\
    \textbf{\model} & {$O(\mathrm{max}({N_1},{N_2})^2)$} \\ \hline
    \label{table:tc}
    \end{tabular}
      \captionof{table}{Time complexity comparison. $N$ and $E$ denote the number of nodes and edges respectively.}
    \end{minipage}
\end{minipage}

Future work will investigate the generation of edit sequences for better interpretability of the predicted similarity, the effects of the usage of other node embedding methods, e.g. GraphSAGE~\cite{hamilton2017inductive}, the adoption of other graph similarity metrics, and the generalization performance of the model across datasets of different domains.

\newpage
\subsubsection*{Acknowledgments}

The work was partially supported by NSF DBI 1565137, NSF DGE1829071, NSF III-1705169, NSF CAREER Award 1741634, NIH U01HG008488, NIH R01GM115833, Snapchat gift funds, and PPDai gift fund.


\bibliographystyle{unsrt}
\bibpunct{[}{]}{,}{n}{}{;} 
\bibliography{bibliography}

\setcounter{section}{0} 

\newpage
\section*{{\Large Appendices}}

\def\thesection{\Alph{section}}
\section{Connections with Set Matching}
\label{sec-theory}

In this section, we present \model from the perspective of set matching, by making theoretical connections with two types of graph matching methods: optimal assignment kernels for graph classification and bipartite graph matching for GED computation. In fact, beyond graphs, set matching has broader applications in Computer Networking (e.g. Internet content delivery)~\cite{maggs2015algorithmic}, Computer Vision (e.g. semantic visual matching)~\cite{zanfir2018deep}, Bioinformatics (e.g. protein alignment)~\cite{zaslavskiy2009global}, Internet Advertising (e.g. advertisement auctions)~\cite{edelman2005advertising}, Labor Markets (e.g. intern host matching)~\cite{roth1984medical}, etc. This opens massive possibilities for future work and suggests the potential impact of \model beyond the graph learning community. 

\subsection{Connection with Optimal Assignment Kernels}

Graph kernels measure the similarity between two graphs, and have been extensively applied to the task of graph classification. Formally speaking, a valid kernel on a set $\mathcal{X}$ is a function $k: \mathcal{X} \times \mathcal{X} \rightarrow \mathbb{R}$ such that there is a real Hilbert space (feature space) $\mathcal{H}$ and a feature map function $\phi: \mathcal{X} \rightarrow \mathbb{R}$ such that $k(x,y) = \langle \phi(x), \phi(y) \rangle$ for every $x$ and $y$ in $\mathcal{X}$, where $\langle \cdot , \cdot \rangle$ denotes the inner product of $\mathcal{H}$. 

Among different families of graph kernels, optimal assignment kernels establish the correspondence between parts of the two graphs, and have many variants~\cite{frohlich2005optimal,johansson2015learning,kriege2016valid,nikolentzos2017matching}. Let $\mathcal{B}(X,Y)$ denote the set of all bijections between two sets of nodes, $X$ and $Y$ Let $k(x,y)$ denote a base kernel that measures the similarity between two nodes $x$ and $y$. An optimal assignment graph kernel $K_{\mathcal{B}}^{k}$ is defined as
\begin{equation}
\begin{aligned}
K_{\mathcal{B}}^{k}(X,Y) = \max_{B \in \mathcal{B}(X,Y)} W(B), \\ \quad \mathrm{where} \quad W(B) = \sum_{(x,y) \in B} k(x,y)
\end{aligned}
\end{equation}
Intuitively, the optimal assignment graph kernels maximize the total similarity between the assigned parts. If the two sets are of different cardinalities, one can add new objects $z$ with $k(z,x)=0, \forall x \in \mathcal{X}$ to the smaller set~\cite{kriege2016valid}.

Let us take the Earth Mover's Distance (EMD) kernel~\cite{nikolentzos2017matching} as an example, since it is among the most similar method to our proposed approach. It treats a graph as a bag of node embedding vectors, but instead of utilizing the pairwise inner products between node embeddings to approximate GED, it computes the optimal ``travel cost'' between two graphs, where the cost is defined as the $L$-2 distance between node embeddings. Given two graphs with node embeddings $\bm{X} \in \mathbb{R}^{N_1 \times D}$ and $\bm{Y} \in \mathbb{R}^{N_2 \times D}$, it solves the following transportation problem~\cite{rubner2000earth}:
\begin{equation}
\label{eq:mne}
\begin{aligned}
\min & \sum_{i=1}^{N_1} \sum_{j=1}^{N_2} \bm{T}_{ij} {||\bm{x}_i - \bm{y}_j||}_{2} \\
& \mathrm{subject \ to} \\
& \sum_{i=1}^{N_1} \bm{T}_{ij} = \frac{1}{N_2} \quad \forall j \in \{ 1,...,N_2 \} \\
& \sum_{j=1}^{N_2} \bm{T}_{ij} = \frac{1}{N_1} \quad \forall i \in \{ 1,...,N_1 \} \\
& \bm{T}_{ij} \geq 0 \quad \forall i \in \{ 1,...,N_1 \}, \forall j \in \{ 1,..,N_2 \}
\end{aligned}
\end{equation}

where $\bm{T} \in \mathbb{R}^{N_1 \times N_2}$ denotes the flow matrix, with $\bm{T}_{ij}$ being how much of node $i$ in $\mathcal{G}_1$ travels (or ``flows'') to node $j$ in $\mathcal{G}_2$. In other words, the EMD between two graphs is the minimum amount of “work” that needs to be done to transform one graph to another, where the optimal transportation plan is encoded by $\bm{T}^{*}$. 

It has been shown that if $N_1=N_2=N$, the optimal solution satisfies $\bm{T}_{ij}^{*} \in \{0,\frac{1}{N}\}$~\cite{balinski1961fixed}, satisfying the optimal bijection requirement of the assignment kernel. Even if $N_1 \neq N_2$, this can still be regarded as approximating an assignment problem~\cite{fan2017point}.

To show the relation between the EMD kernel and our approach, we consider \model as a mapping function that, given two graphs with node embeddings $\bm{X} \in \mathbb{R}^{N_1 \times D}$ and $\bm{Y} \in \mathbb{R}^{N_2 \times D}$, produces one score as the predicted similarity score, which is compared against the ground-truth similarity score:
\begin{equation}
\label{eq:mne_equivalent}
\begin{aligned}
\min (h_{\Theta} (\bm{X}, \bm{Y}) - \lambda(\mathrm{GED}(\mathcal{G}_1, \mathcal{G}_2)))^2
\end{aligned}
\end{equation}
where $h_{\Theta} (\bm{X}, \bm{Y})$ represents the Graph Interaction and CNN layers but can potentially be replaced by any neural network transformation.

To further see the connection, we consider one CNN layer with one filter of size $N$ by $N$, where $N = \mathrm{max}(N_1, N_2)$. Then Eq.~\ref{eq:mne_equivalent} becomes:
\begin{equation}
\begin{aligned}
\min (\sigma (\sum_{i=1}^{N} \sum_{j=1}^{N} \bm{\Theta}_{ij} (\bm{x}_i^{T} \bm{y}_j)) - \lambda(\mathrm{GED}(\mathcal{G}_1, \mathcal{G}_2)))^2
\end{aligned}
\end{equation}
where $\bm{\Theta} \in \mathbb{R}^{N \times N}$ is the convolutional filter.

Compared with the EMD kernel, our method has two benefits. (1) The mapping function and the node embeddings $\bm{X}$ and $\bm{Y}$ are simultaneously learned through backpropagation, while the kernel method solves the assignment problem to obtain $\bm{T}$ and uses fixed node embeddings $\bm{X}$ and $\bm{Y}$, e.g. generated by the decomposition of the graph Laplacian matrix. Thus, \model is suitable for \textit{learning} an approximation of the GED graph distance metric, while the kernel method cannot. The typical usage of a graph kernel is to feed the graph-graph similarities into a SVM classifier for graph classification. (2) The best average time complexity of solving Eq.~\ref{eq:mne} scales $O(N^3 log N)$~\cite{pele2009fast}, where $N$ denotes the
number of total nodes in two graphs, while the convolution operation is in O($(\mathrm{max}(N_1,N_2))^2$) time.

\subsection{Connection with Bipartite Graph Matching}
\label{subsec-bgm}

Among the existing approximate GED computation algorithms, Hungarian~\cite{riesen2009approximate} and VJ~\cite{fankhauser2011speeding} are two classic ones based on bipartite graph matching. Similar to the optimal assignment kernels, Hungarian and VJ also find an optimal match between the nodes of two graphs. However, different from the EMD kernel, the assignment problem has stricter constraints: One node in $\mathcal{G}_1$ can be only mapped to one other node in $\mathcal{G}_2$. Thus, the entries in the assignment matrix $\bm{T} \in \mathbb{R}^{N^{'} \times N^{'}}$ are either 0 or 1, denoting the operations transforming $\mathcal{G}_1$ into $\mathcal{G}_2$, where $N^{'} = N_1 + N_2$. The assignment problem takes the following form:
\begin{equation}
\label{eq:bgm}
\begin{aligned}
\min & \sum_{i=1}^{N^{'}} \sum_{j=1}^{N^{'}} \bm{T}_{ij} \bm{C}_{ij} \\
& \mathrm{subject \ to} \\
& \sum_{i=1}^{N^{'}} \bm{T}_{ij} = 1 \quad \forall j \in \{ 1,...,N^{'} \} \\
& \sum_{j=1}^{N^{'}} \bm{T}_{ij} = 1 \quad \forall i \in \{ 1,...,N^{'} \} \\
& \bm{T}_{ij} \in \{0, 1\} \quad \forall i \in \{ 1,...,N^{'} \}, \forall j \in \{ 1,..,N^{'} \}
\end{aligned}
\end{equation}
The cost matrix $\bm{C} \in \mathbb{R}^{N^{'} \times N^{'}}$ reflects the GED model, and is defined as follows:
\[
\renewcommand\arraystretch{1.3}
\bm{C} = \mleft[
\begin{array}{ccc|ccc}
  \bm{C}_{1,1} & \dots & \bm{C}_{1,N_2} & \bm{C}_{1,\epsilon} & \dots & \infty \\
  \vdots & \ddots & \vdots & \vdots & \ddots & \vdots \\
  \bm{C}_{N_1,1} & \dots & \bm{C}_{N_1,N_2} & \infty & \dots & \bm{C}_{N_1,\epsilon} \\
  \hline
  \bm{C}_{\epsilon,1} & \dots & \infty & 0 & \dots & 0 \\
  \vdots & \ddots & \vdots & \vdots & \ddots & \vdots \\
  \infty & \dots & \bm{C}_{\epsilon,N_2} & 0 & \dots & 0 \\
\end{array}
\mright]
\]
where $\bm{C}_{i,j}$ denotes the cost of a substitution, $\bm{C}_{i,\epsilon}$ denotes the cost of a node deletion, and $\bm{C}_{\epsilon,j}$ denotes the cost of a node insertion. According to our GED definition, $\bm{C}_{ij} = 0$ if the labels of node $i$ and node $j$ are the same, and 1 otherwise; $\bm{C}_{i,\epsilon} = \bm{C}_{\epsilon,j} = 1$.

Exactly solving this constrained optimization program would yield the exact GED solution~\cite{fankhauser2011speeding}, but it is NP-complete since it is equivalent to finding an optimal matching in a complete bipartite graph~\cite{riesen2009approximate}. 


To efficiently solve the assignment problem, the Hungarian algorithm~\cite{kuhn1955hungarian} and the Volgenant Jonker (VJ)~\cite{jonker1987shortest} algorithm are commonly used, which both run in cubic time. In contrast, \model takes advantage of the exact solutions of this problem during the training stage, and computes the approximate GED during testing in quadratic time, without the need for solving any optimization problem for a new graph pair.

\subsection{Summary of Connections with Set Matching}

To sum up, our model, \model, represents a new approach to modeling the similarities between graphs, by first transforming each graph into a set of node embeddings, where embeddings encode the link structure around each node, and then matching two sets of node embeddings. The entire model can be trained in an end-to-end fashion. In contrast, the other two approaches in Table~\ref{table:set_matching_summary} also model the graph-graph similarity by viewing a graph as a set, but suffer from limited learnability and cannot be trained end-to-end. Due to its neural network nature, the convolutional set matching approach enjoys flexibility and thus has the potential to be extended to solve other set matching problems.

\begin{table}[H]
\caption{Summary of three set matching based approaches to graph similarity. $f(\cdot)$ denotes the graph transforming function, and $g(\cdot,\cdot)$ denotes the set matching function. They are defined in Section 2 in the main paper.}
\small
\begin{tabular}
{cccc} \hline
\textbf{Approach} & \textbf{Example(s)} & \textbf{$f(\mathcal{G})$} & \textbf{$g(\mathcal{G}_1,\mathcal{G}_2)$} \\ \hline
\textbf{Optimal Alignment Kernels} & EMD kernel~\cite{nikolentzos2017matching} & Node Embedding & Solver of Eq.~\ref{eq:mne} \\
\textbf{Bipartite Graph Matching} & Hungarian~\cite{riesen2009approximate}, VJ~\cite{fankhauser2011speeding} & Nodes of $\mathcal{G}$ & Solver of Eq.~\ref{eq:bgm} \\
\textbf{Convolutional Set Matching} & \model & Node Embedding & Graph Interaction + CNNs \\ \hline
\end{tabular}
\centering
\label{table:set_matching_summary}
\end{table}

\section{Dataset Description}

Three real-world graph datasets are used for the experiments. A concise summary can be found in Table~\ref{table:dataset_summary}.

\textbf{AIDS} \enspace AIDS is a collection of antivirus screen chemical compounds from the Developmental Therapeutics Program at NCI/NIH 7~\footnote{\url{https://wiki.nci.nih.gov/display/NCIDTPdata/AIDS+Antiviral+Screen+Data}.}, and has been used in several existing work on graph similarity search~\cite{zeng2009comparing,wang2012efficient}. It contains 42,687 chemical compound structures with Hydrogen atoms omitted. We randomly select 700 graphs, each of which has 10 or less than 10 nodes.

\textbf{LINUX} \enspace The LINUX dataset was originally introduced in \cite{wang2012efficient}. It consists of 48,747 Program Dependence Graphs (PDG) generated from the Linux kernel. Each graph represents a function, where a node represents one statement and an edge represents the dependency between the two statements. We randomly select 1000 graphs of equal or less than 10 nodes each.

\textbf{IMDB} \enspace The IMDB dataset~\cite{yanardag2015deep} (named ``IMDB-MULTI'') consists of 1500 ego-networks of movie actors/actresses, where there is an edge if the two people appear in the same movie. To test the scalability and efficiency of our proposed approach, we use the full dataset without any selection.

Since the GED computation is pairwise, it is necessary to take the number of pairs into consideration. There are 294K, 0.6M and 1.35M total number of training graph pairs in the AIDS, LINUX and IMDB dataset, respectively.

\begin{table}[H]
\caption{Statistics of datasets. ``Min'', ``Max'', ``Mean'', and ``Std'' refer to the minimum, maximum, mean, and standard deviation of the graph sizes (number of nodes), respectively.}
\small
\begin{tabular}
{cccccccc} \hline
\textbf{Dataset} & \textbf{Meaning} & \textbf{\#Node Labels} & \textbf{\#Graphs} & \textbf{Min} & \textbf{Max} & \textbf{Mean} & \textbf{Std}\\ \hline
\textbf{AIDS} & Chemical Compounds & 29 & 700 & 2 & 10 & 8.9 & 1.4 \\
\textbf{LINUX} & Program Dependence Graphs & 1 & 1000 & 4 & 10 & 7.7 & 1.5  \\
\textbf{IMDB} & Ego-Networks & 1 & 1500 & 7 & 89 & 13.0 & 8.5 \\ \hline
\end{tabular}
\centering
\label{table:dataset_summary}
\end{table}

\section{Baseline Details}
Our baselines include two types of approaches, fast approximate GED computation algorithms and neural network based models. 

The first category of baselines includes four classic algorithms for GED computation. (1) \textit{A*-Beamsearch (Beam)}, (2) \textit{Hungarian}, and (3) \textit{VJ} return upper bounds of the true GEDs. (2) and (3) are described in Section~\ref{subsec-bgm}. (4) \textit{HED}~\cite{fischer2015approximation} is based on Hausdorff matching, and yields GEDs smaller than or equal to the actual GEDs. Therefore, Beam, Hungarian, and VJ are used to determine the ground truth for IMDB without considering HED.

The second category of baselines includes the following neural network architectures. (1) \textit{Siamese MPNN}~\cite{ribalearning}, (2) \textit{EmbAvg} , (3) \textit{GCNMean} and (4) \textit{GCNMax}~\cite{defferrard2016convolutional} are four neural network architectures. (1) generates all the pairwise node embedding similarity scores, and for each node, it finds one node in the other graph with the highest similarity score. It simply sums up all these similarity scores as the final result. (2), (3), and (4) take the dot product of the graph-level embeddings of the two graphs to produce the similarity score. (2) takes the unweighted average of node embeddings as the graph embedding. (3) and (4) adopt the original GCN architectures based on graph coarsening with mean and max pooling, respectively, to gain the graph embedding. \textit{\model} is our complete model with three levels of comparison granularities.
\section{Parameter Setting and Implementation Details} 

For the proposed model, we use the same network architecture on all the datasets. We set the number of GCN layers to 3, and use ReLU as the activation function. For the resizing scheme, all the similarity matrices are resized to 10 by 10. For the CNNs, we use the following architecture: conv(6,1,1,16), maxpool(2), conv(6,1,16,32), maxpool(2), conv(5,1,32,64), maxpool(2), conv(5,1,64,128), maxpool(3), conv(5,1,128,128), maxpool(3) (``conv(window size, kernel stride, input channels, output channels)''; ``maxpool(pooling size)''). 

\model is implemented using TensorFlow, and tested on a single machine with an Intel i7-6800K CPU and one Nvidia Titan GPU. As for training, we set the batch size to 128, use the Adam algorithm for optimization~\cite{kingma2014adam}, and fix the initial learning rate to 0.001. We set the number of iterations to 15000, and select the best model based on the lowest validation loss. 

\section{Discussion and Result Analysis} 

\subsection{Comparison between \model and Baselines}

The classic algorithms for GED computation (e.g. A*, Beam, Hungarian, VJ, HED, etc.) usually require rather complicated design and implementation based on discrete optimization or combinatorial search. In contrast, \model is learnable and can take advantage of the exact solutions of GED computation during the training stage. Regarding time complexity, \model computes the approximate GED in quadratic time, without the need for solving any optimization problem for a new graph pair. In fact, \model computes the similarity score. However, it can be mapped back to the corresponding GED.

For the simple neural network based approaches:

1. For EmbAvg, GCNMean and GCNMax, they are all calculating the similarity score based on the inner product of graph embeddings. For EmbAvg, it first takes the unweighted average of node embeddings to gain the graph embedding, while GCNMean and GCNMax adopt graph coarsening with mean and max pooling, respectively, to obtain the graph embedding. The potential issue for these models are: (1) They fail to leverage the information from fine-grained node-level comparisons; (2) They calculate the final result based on the inner product between two graph embeddings, without any module to learn the graph level interactions. In contrast, \model constructs multi-scale similarity matrices to encode the node-level interaction at different scales, and adopts CNNs to detect the matching patterns afterwards.

2. For Siamese MPNN, it also computes all the pairwise node embedding similarity scores, like \model. However, it goes through all the nodes in both graphs, and for every node, it finds one node in the other graph with the highest similarity score, and simply sums up all these similarity scores as the final result with no trainable components afterwards. Our model, instead, is equipped with the learnable CNN kernels and dense layers to extract matching patterns with similarity scores from multiple scales.


For efficiency, notice that A* can no longer be used to provide the ground truth for IMDB, the largest dataset, as ``no currently available algorithm manages to reliably compute GED within reasonable time between graphs with more than 16 nodes''~\cite{blumenthal2018exact}. This not only shows the significance of time reduction for computing pairwise graph similarities/distances, but also highlights the challenges of creating a fast and accurate graph similarity search system. 

As seen in Table 1 and Fig. 4 in the main paper, \model strikes a good balance between effectiveness and efficiency. Specifically, \model achieves the smallest error, the best ranking performance, and great time reduction on the task of graph similarity search. 

\subsection{Comparison between \model and Simple Variants of \model}
\begin{table}[H]
\small
\centering
\vspace{-0.05in}
\caption{\model-L1-Pad and \model-L1-Resize use the node embedding by the last of the three GCN layers to generate the similarity matrix, with the padding and resizing schemes, respectively.}
  \begin{tabular}{cccccccccc}
    \toprule
    \multirow{3}{*}{\textbf{Method}} &
      \multicolumn{3}{c}{\textbf{AIDS}} &
      \multicolumn{3}{c}{\textbf{LINUX}} &
      \multicolumn{3}{c}{\textbf{IMDB}} \\
      & {mse} & {$\tau$} & {p@10} & {mse} & {$\tau$} & {p@10} & {mse} & {$\tau$} & {p@10}\\
      \midrule
    \model-L1-Pad, & 0.807 & 0.714 & 0.514 & 0.141 & 0.951 & 0.983 & 6.455 & 0.562 & 0.552 \\
    \model-L1-Resize & 0.811 & 0.715 & 0.499 & 0.103 & 0.959 & 0.986 & 0.896 & 0.828 & 0.810 \\
    \midrule
    \model & \textbf{0.787} & \textbf{0.724} & \textbf{0.534} & \textbf{0.058} & \textbf{0.962} & \textbf{0.992} & \textbf{0.743} & \textbf{0.847} & \textbf{0.828} \\
    \bottomrule
  \end{tabular}
  \vspace{-1.8em}
\centering
\label{table:extra_results}
\end{table}

The effectiveness of multiple levels of comparison over a single level can be seen from the performance boost from the last two rows of Table~\ref{table:extra_results}. The improvement is especially significant on IMDB, which can be attributed to the large average graph size, as seen from Table~\ref{table:dataset_summary}. The large variance of graph sizes associated with IMDB also favors the proposed resizing scheme over the padding scheme, which is also reflected in the results. On AIDS and LINUX, the use of resizing does not improve the performance much, but on IMDB, the improvement is much more significant.

\section{Extra Visualization} 

\begin{figure}
    \centering
    \subfloat[Query ranking result.]
    {{\includegraphics[scale=0.57]{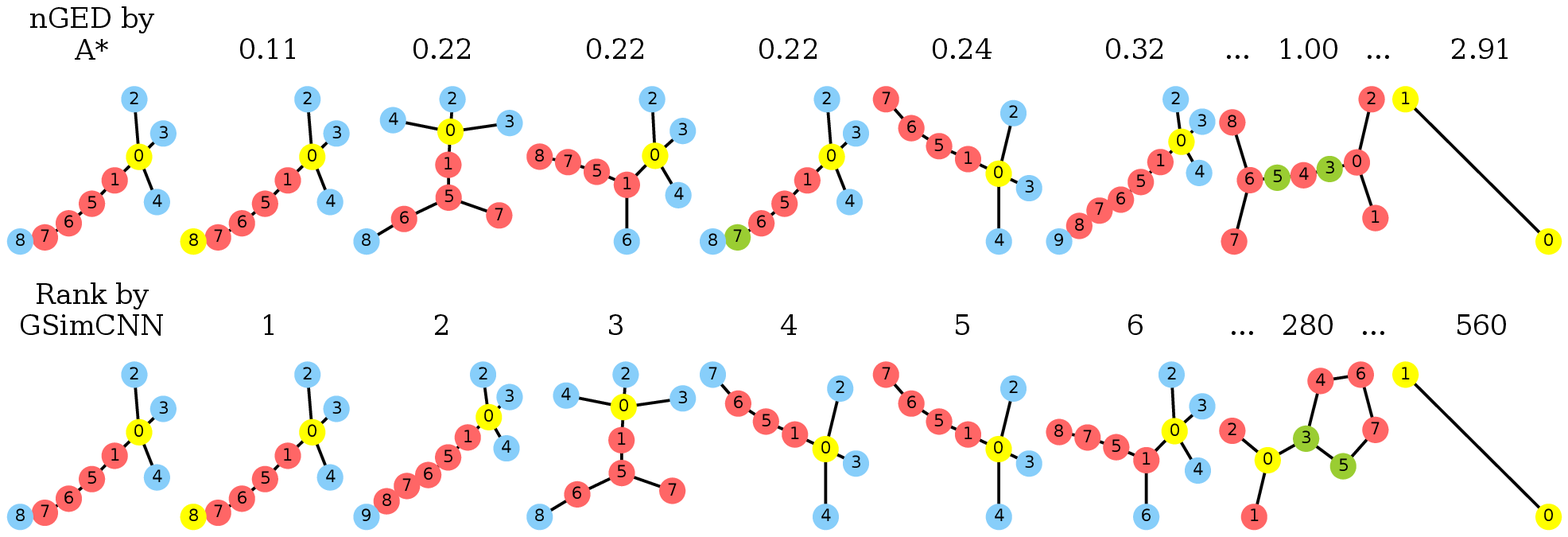}}}
    \vspace{-0.0in}
    \centering
    \subfloat[Similarity matrix between the query and the most similar graph ranked by both methods.]
    {{\includegraphics[scale=0.38]{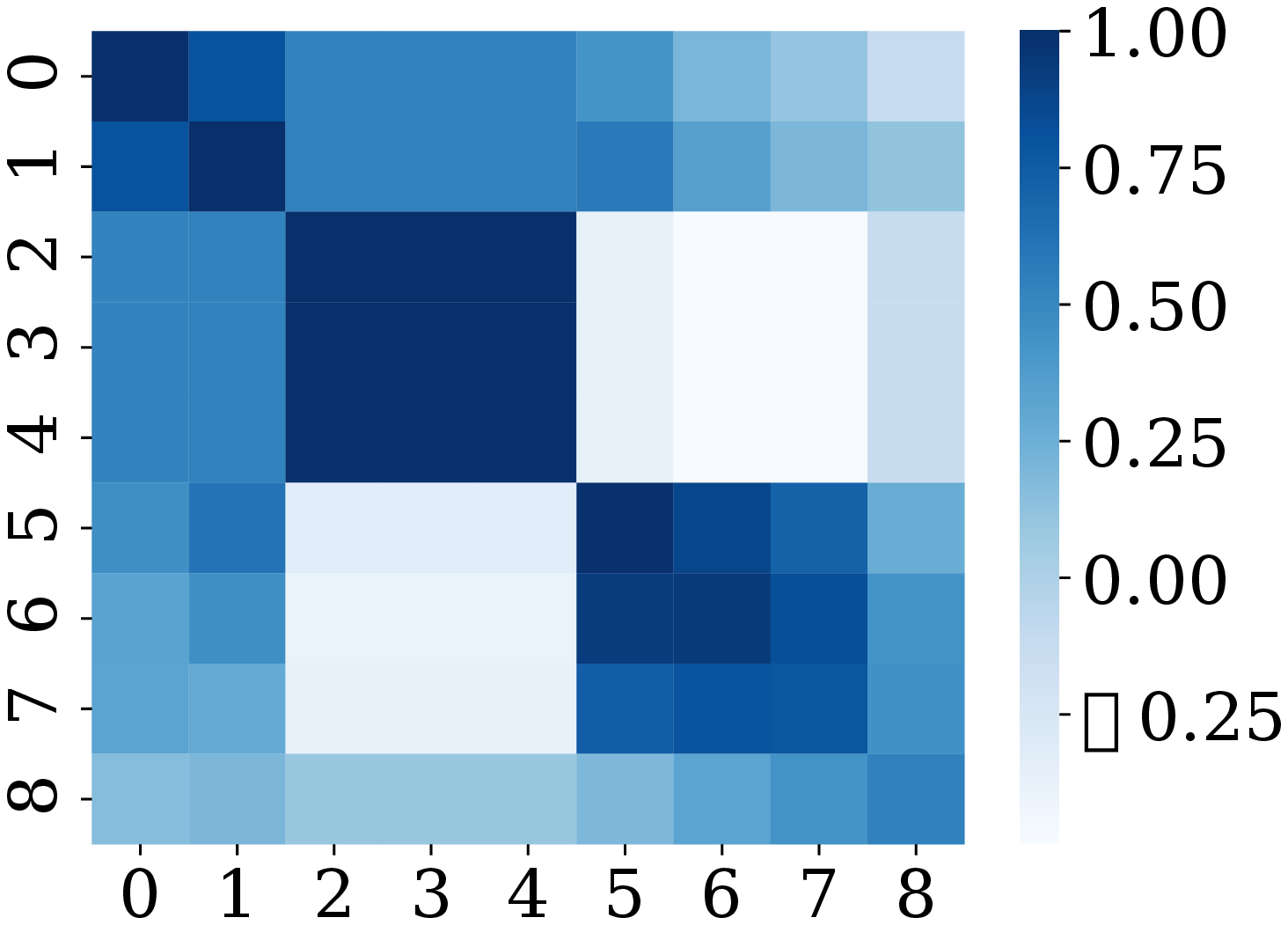}}}
    \vspace{-0.0in}
    \centering
    \subfloat[Similarity matrix between the query and the least similar graph ranked by both methods.]
    {{\includegraphics[scale=0.38]{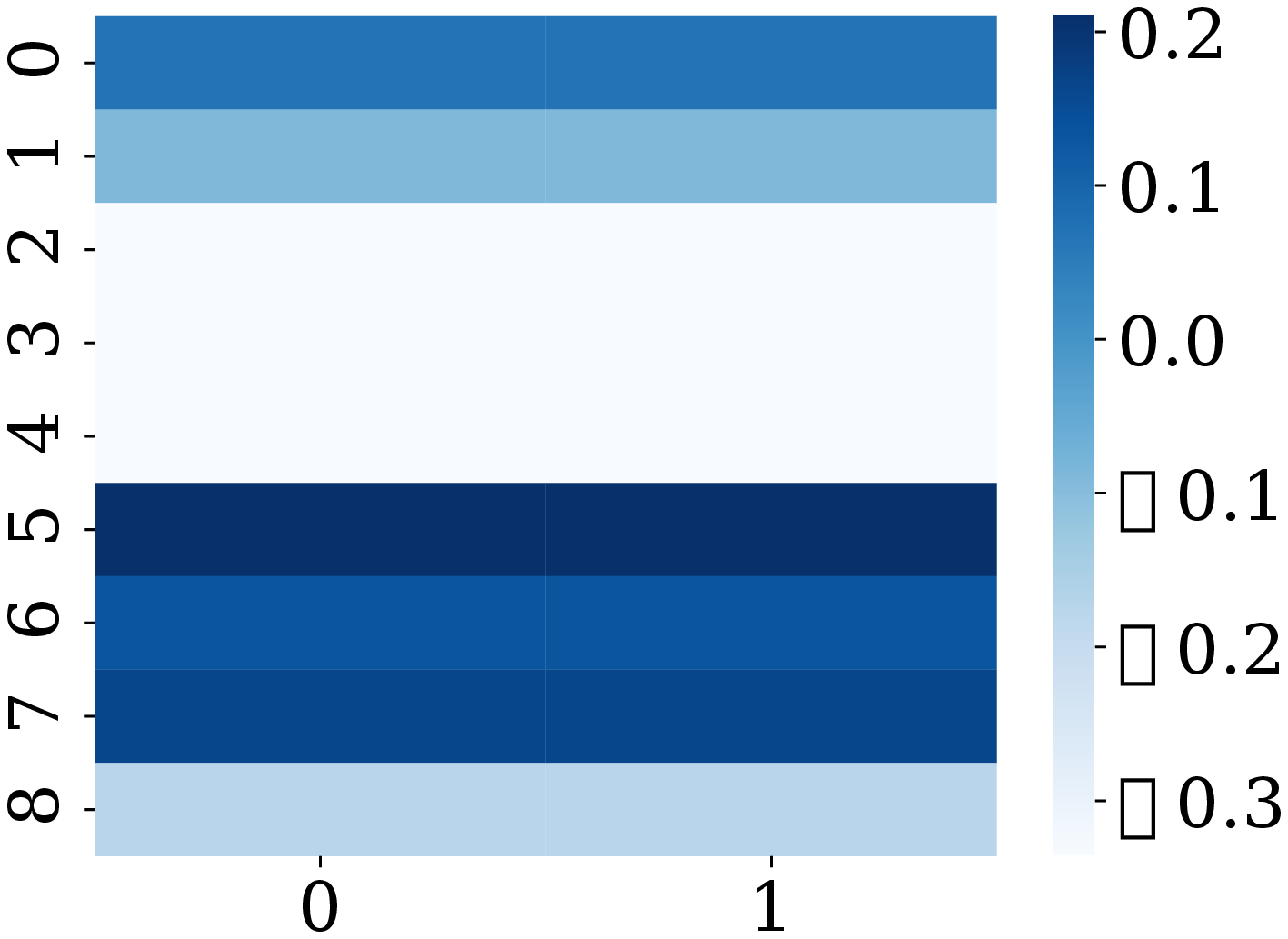}}}
    \vspace{-0.0in}
    \caption{A query case study on AIDS.}
    \label{fig:search_demo_extra_1}
\end{figure}

\begin{figure}[H]
    \centering
    \subfloat[Query ranking result.]
    {{\includegraphics[scale=0.57]{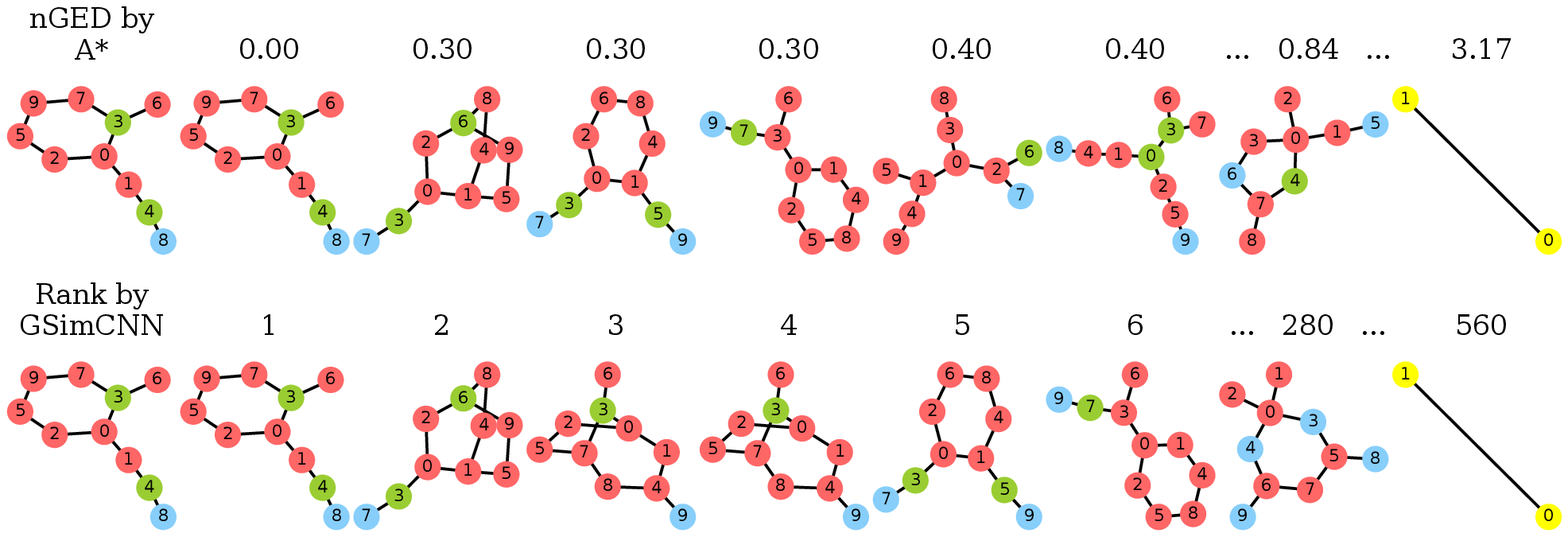}}}
    \vspace{-0.0in}
    \centering
    \subfloat[Similarity matrix between the query and the most similar graph ranked by both methods.]
    {{\includegraphics[scale=0.38]{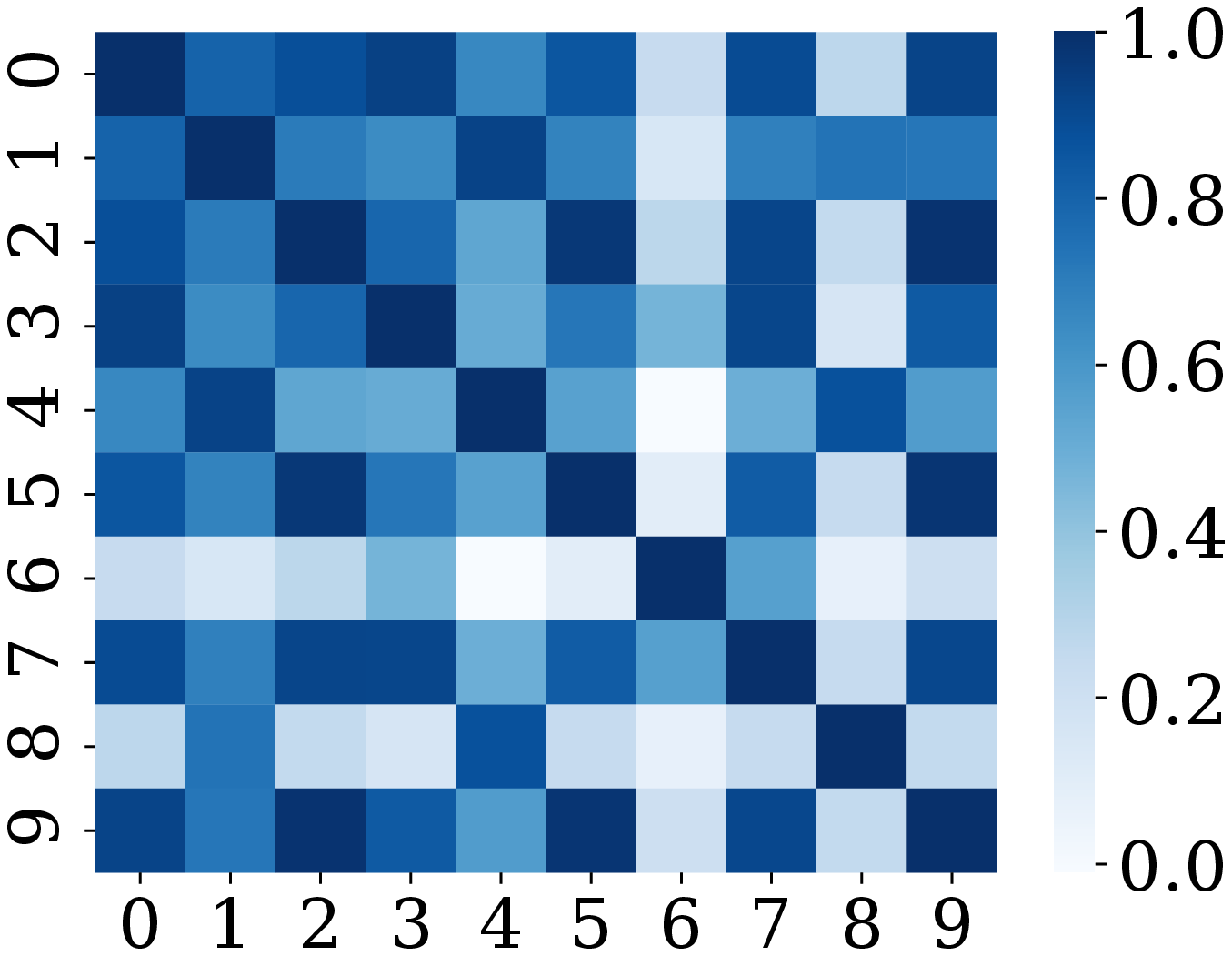}}}
    \vspace{-0.0in}
    \centering
    \subfloat[Similarity matrix between the query and the least similar graph ranked by both methods.]
    {{\includegraphics[scale=0.38]{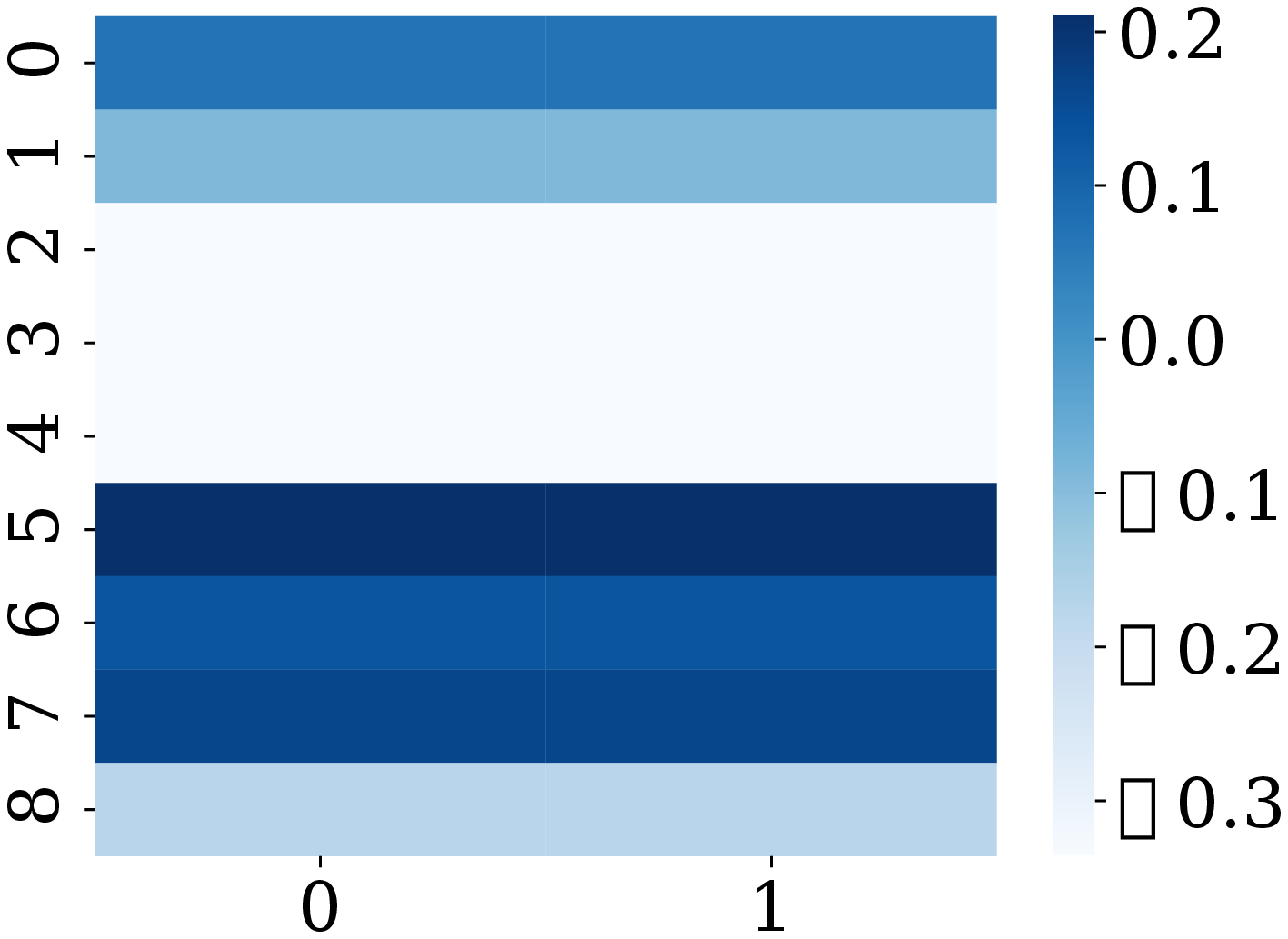}}}
    \vspace{-0.0in}
    \caption{A query case study on AIDS.}
    \label{fig:search_demo_extra_2}
\end{figure}

\begin{figure}
    \centering
    \subfloat[Query ranking result.]
    {{\includegraphics[scale=0.57]{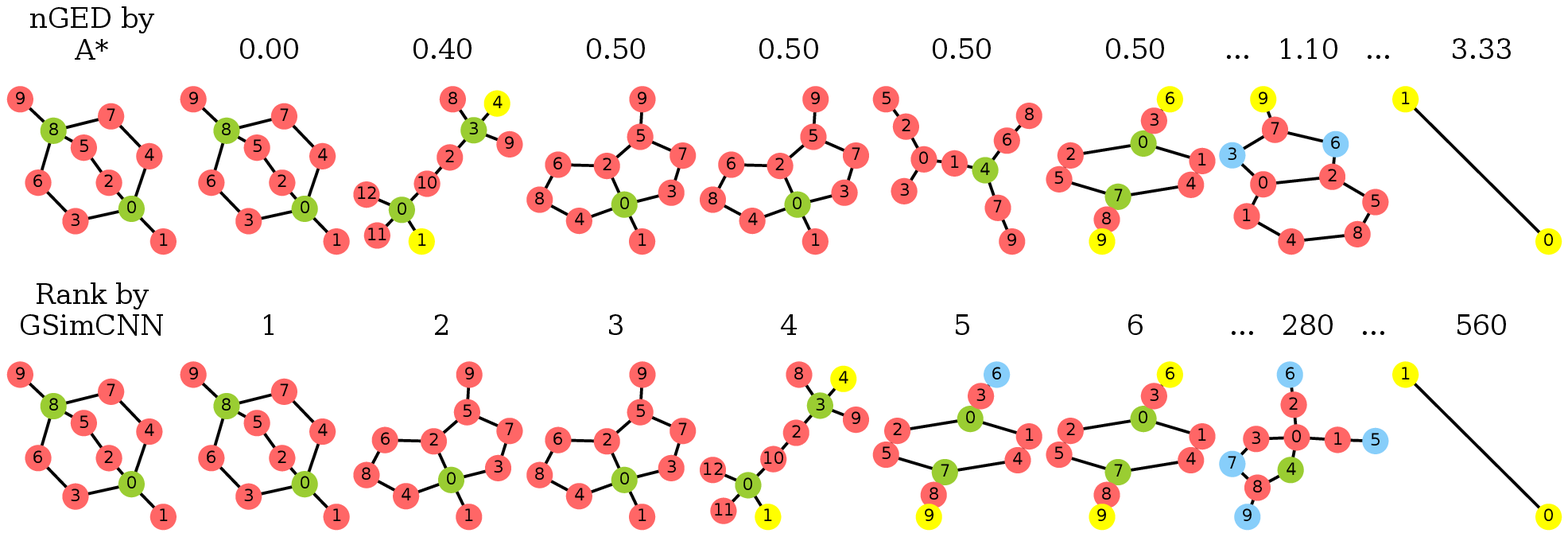}}}
    \vspace{-0.0in}
    \centering
    \subfloat[Similarity matrix between the query and the most similar graph ranked by both methods.]
    {{\includegraphics[scale=0.38]{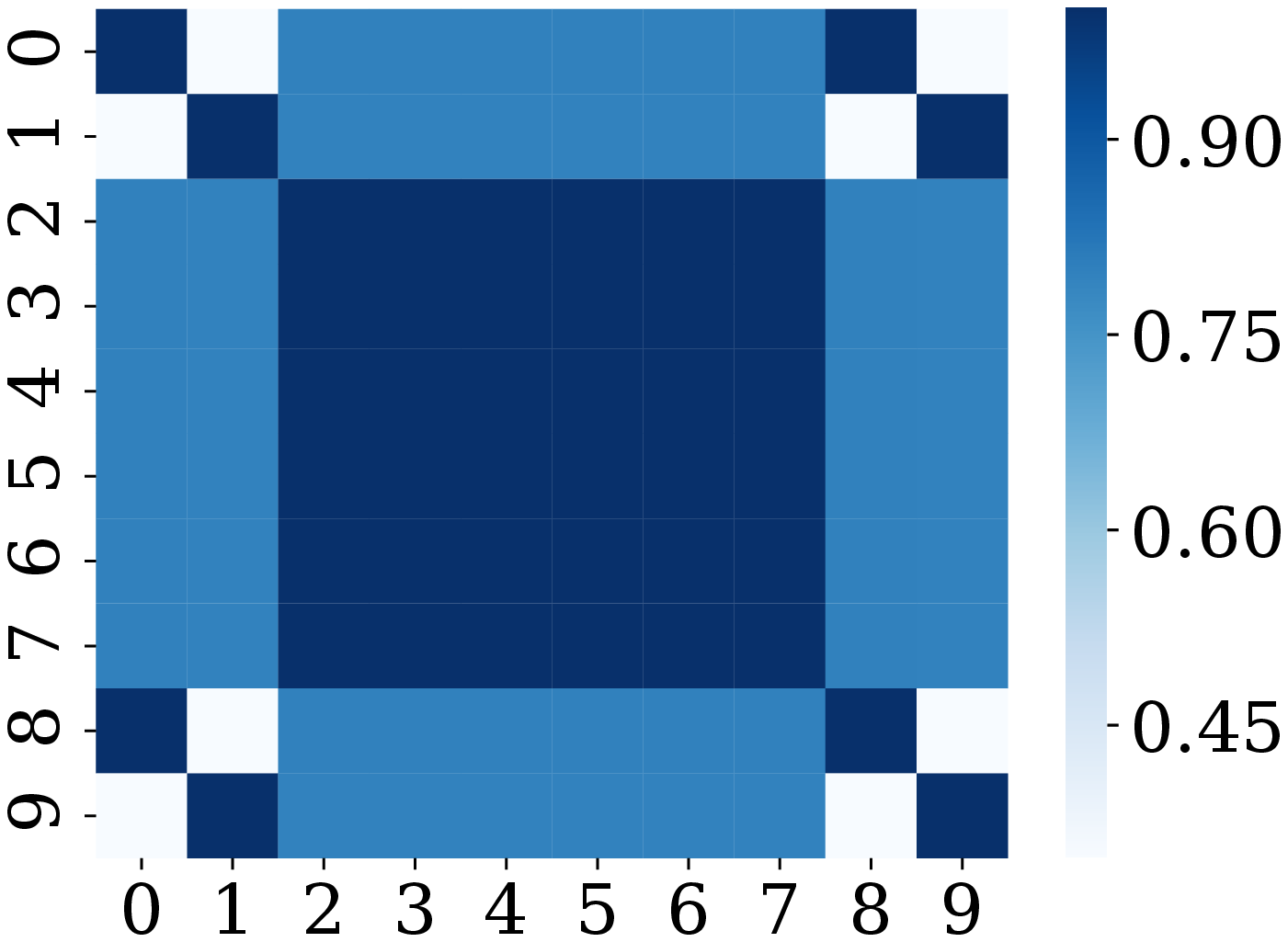}}}
    \vspace{-0.0in}
    \centering
    \subfloat[Similarity matrix between the query and the least similar graph ranked by both methods.]
    {{\includegraphics[scale=0.38]{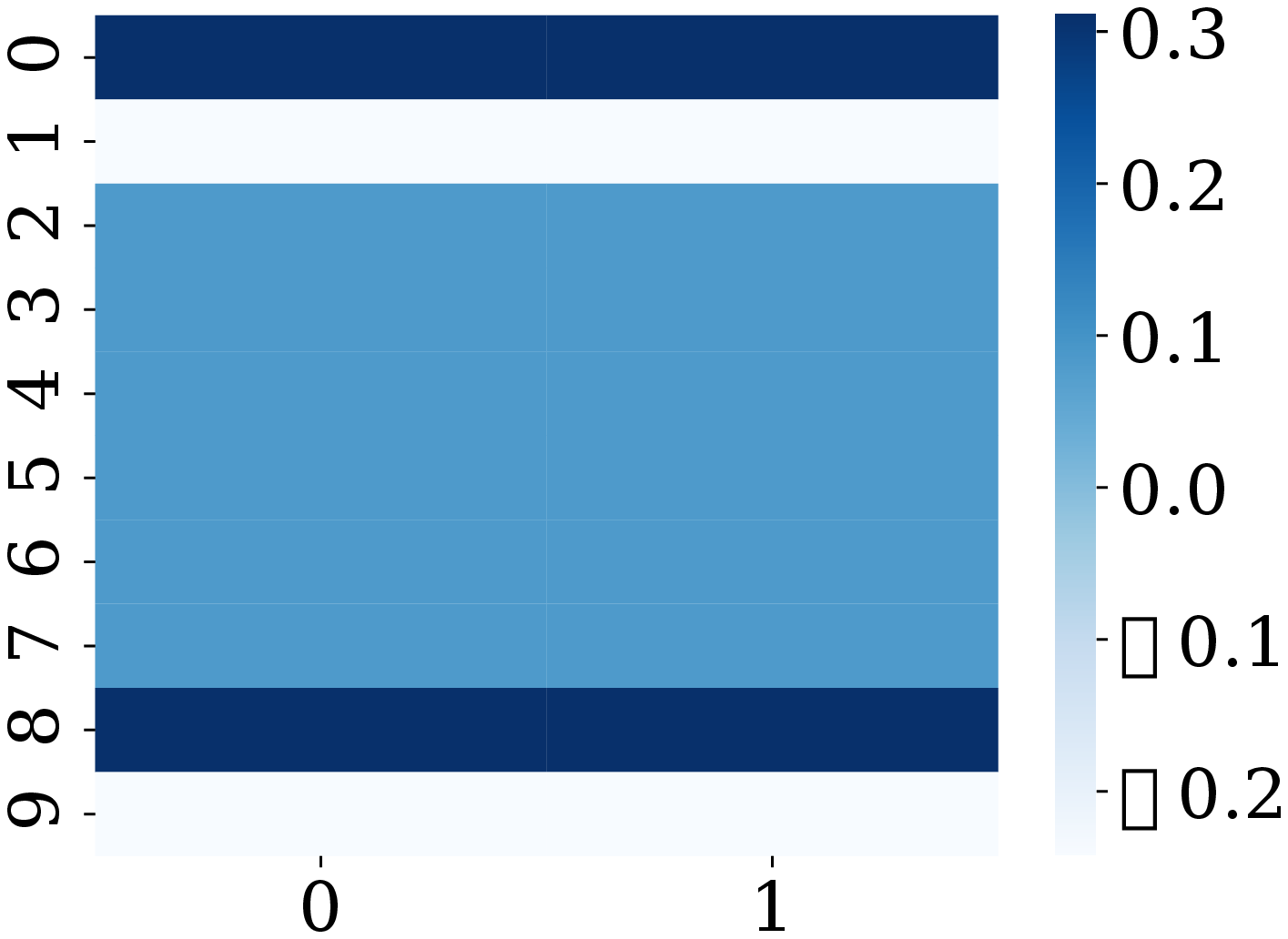}}}
    \vspace{-0.0in}
    \caption{A query case study on AIDS.}
    \label{fig:search_demo_extra_3}
\end{figure}

\begin{figure}
    \centering
    \subfloat[Query ranking result.]
    {{\includegraphics[scale=0.57]{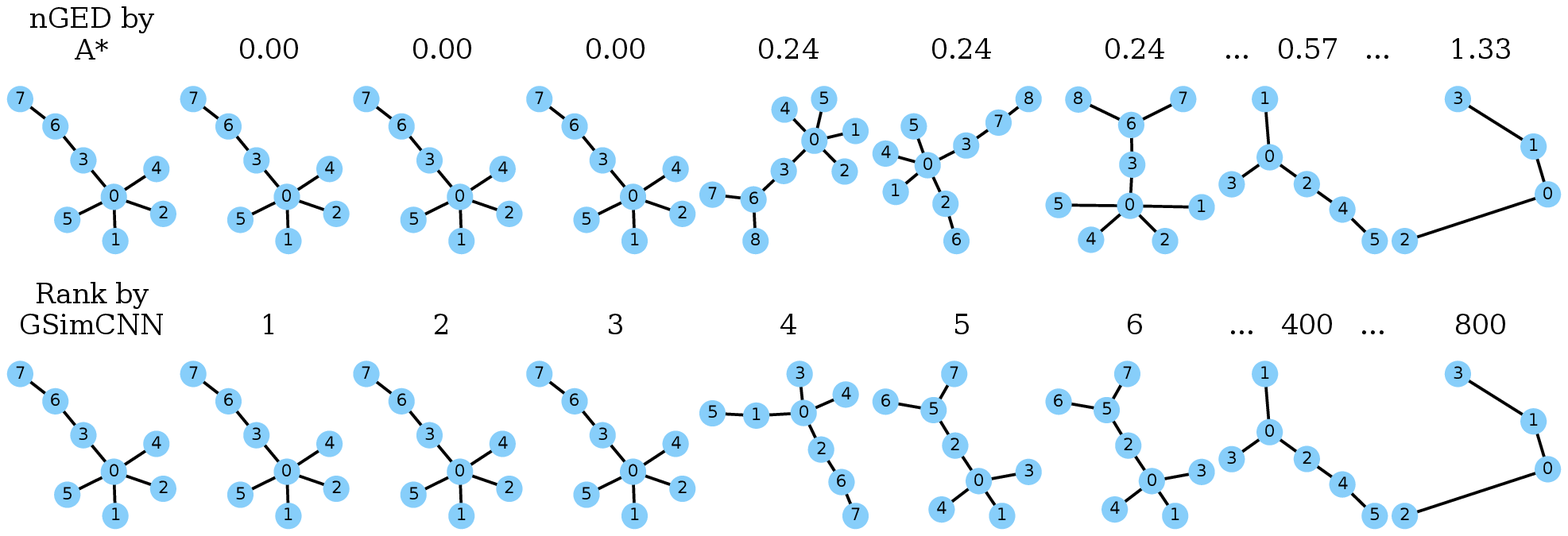}}}
    \vspace{-0.0in}
    \centering
    \subfloat[Similarity matrix between the query and the most similar graph ranked by both methods.]
    {{\includegraphics[scale=0.38]{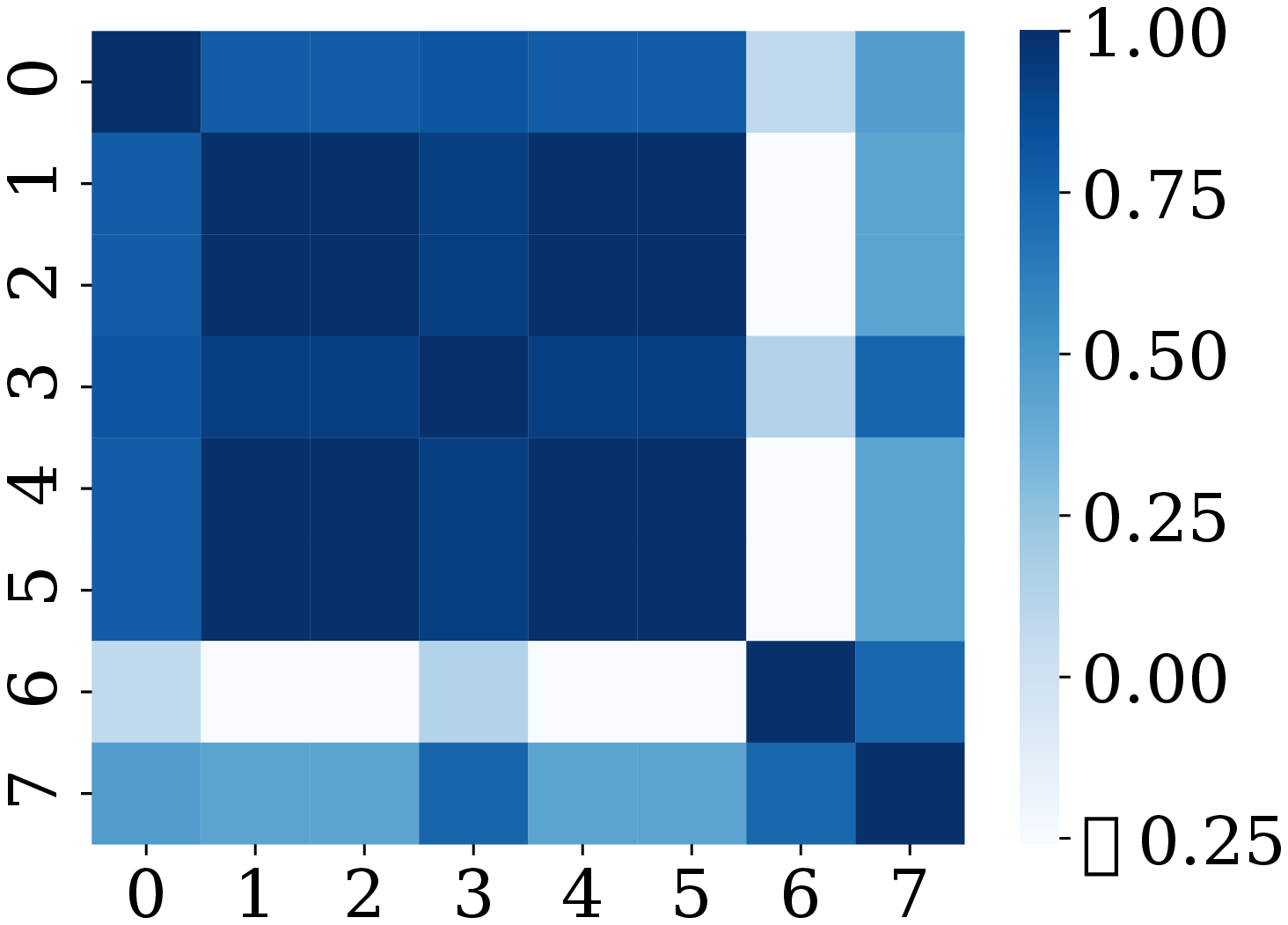}}}
    \vspace{-0.0in}
    \centering
    \subfloat[Similarity matrix between the query and the least similar graph ranked by both methods.]
    {{\includegraphics[scale=0.38]{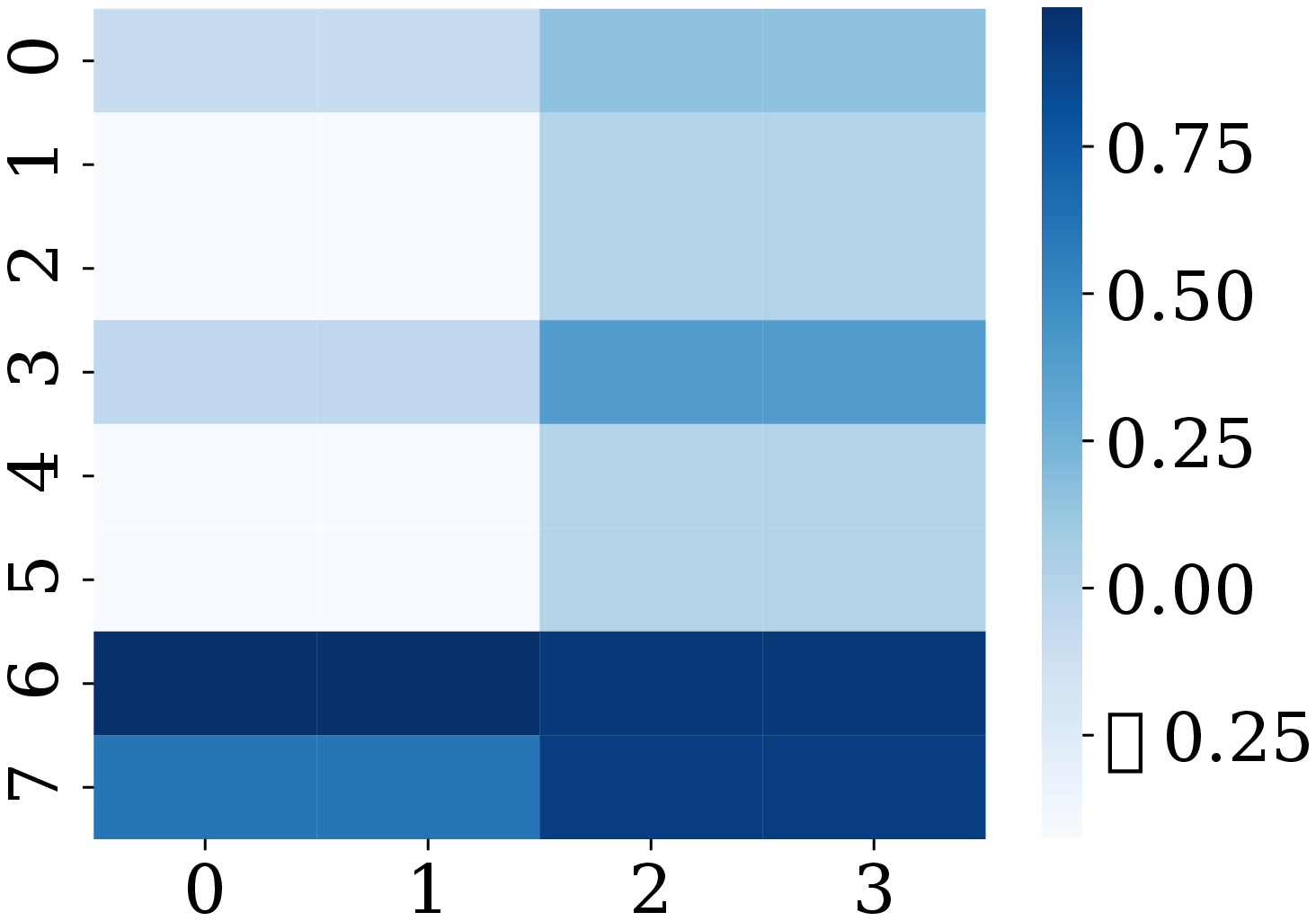}}}
    \vspace{-0.0in}
    \caption{A query case study on LINUX.}
    \label{fig:search_demo_extra_4}
\end{figure}

\begin{figure}
    \centering
    \subfloat[Query ranking result.]
    {{\includegraphics[scale=0.57]{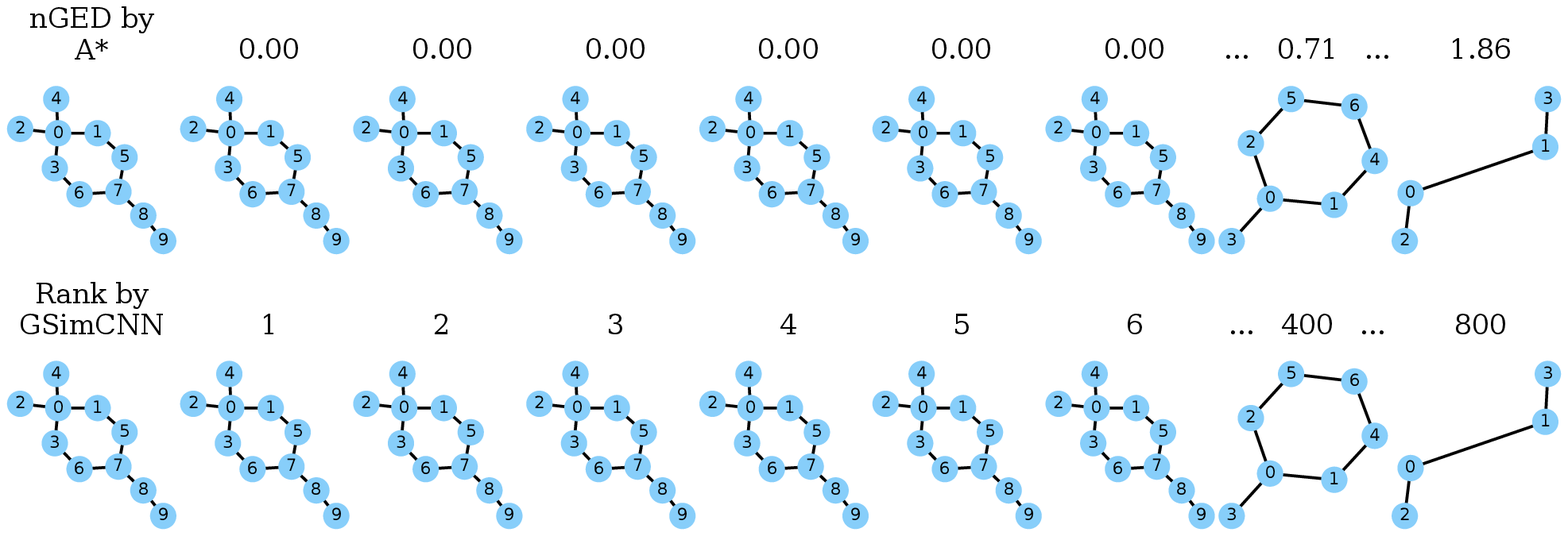}}}
    \vspace{-0.0in}
    \centering
    \subfloat[Similarity matrix between the query and the most similar graph ranked by both methods.]
    {{\includegraphics[scale=0.38]{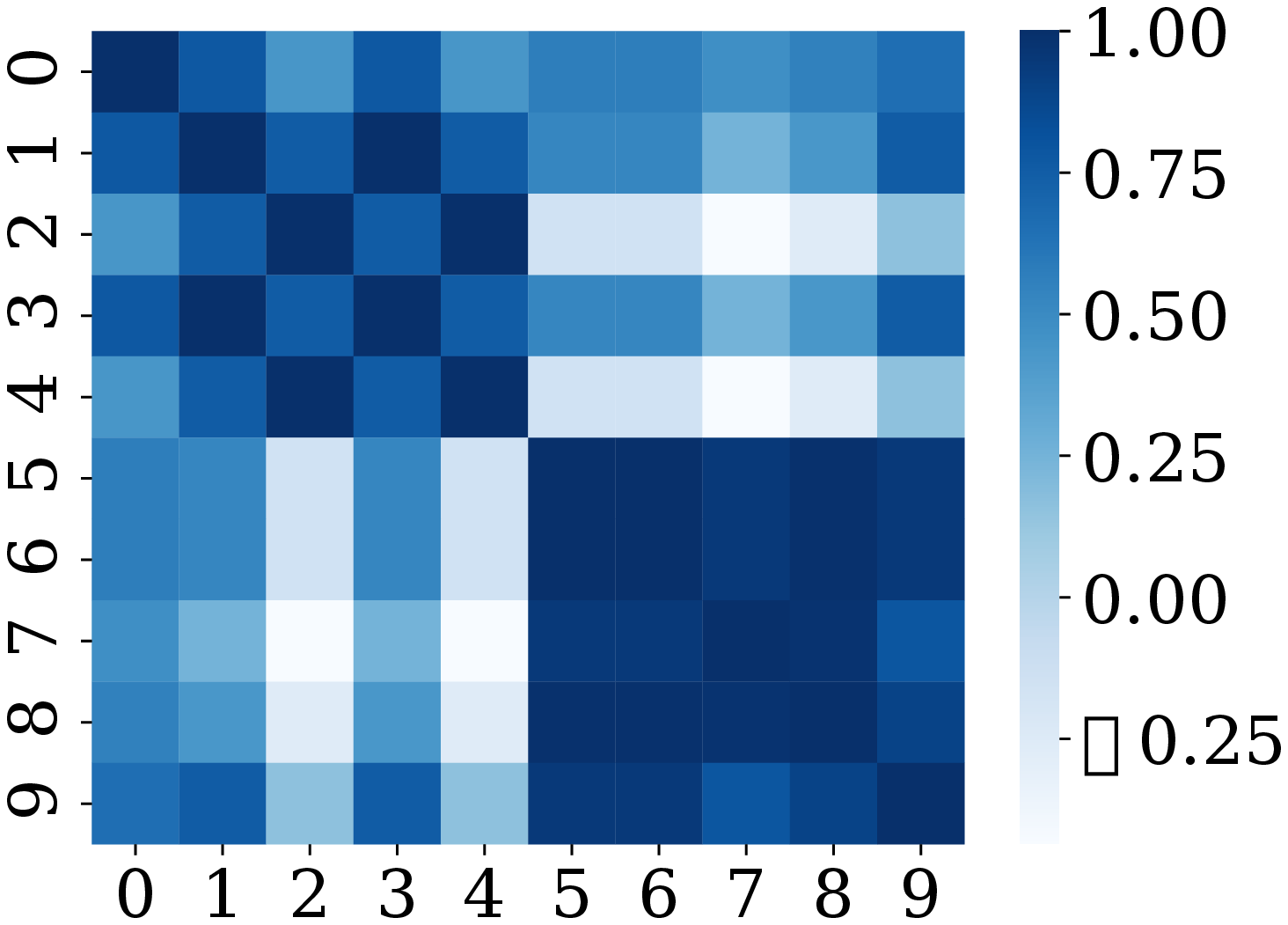}}}
    \vspace{-0.0in}
    \centering
    \subfloat[Similarity matrix between the query and the least similar graph ranked by both methods.]
    {{\includegraphics[scale=0.38]{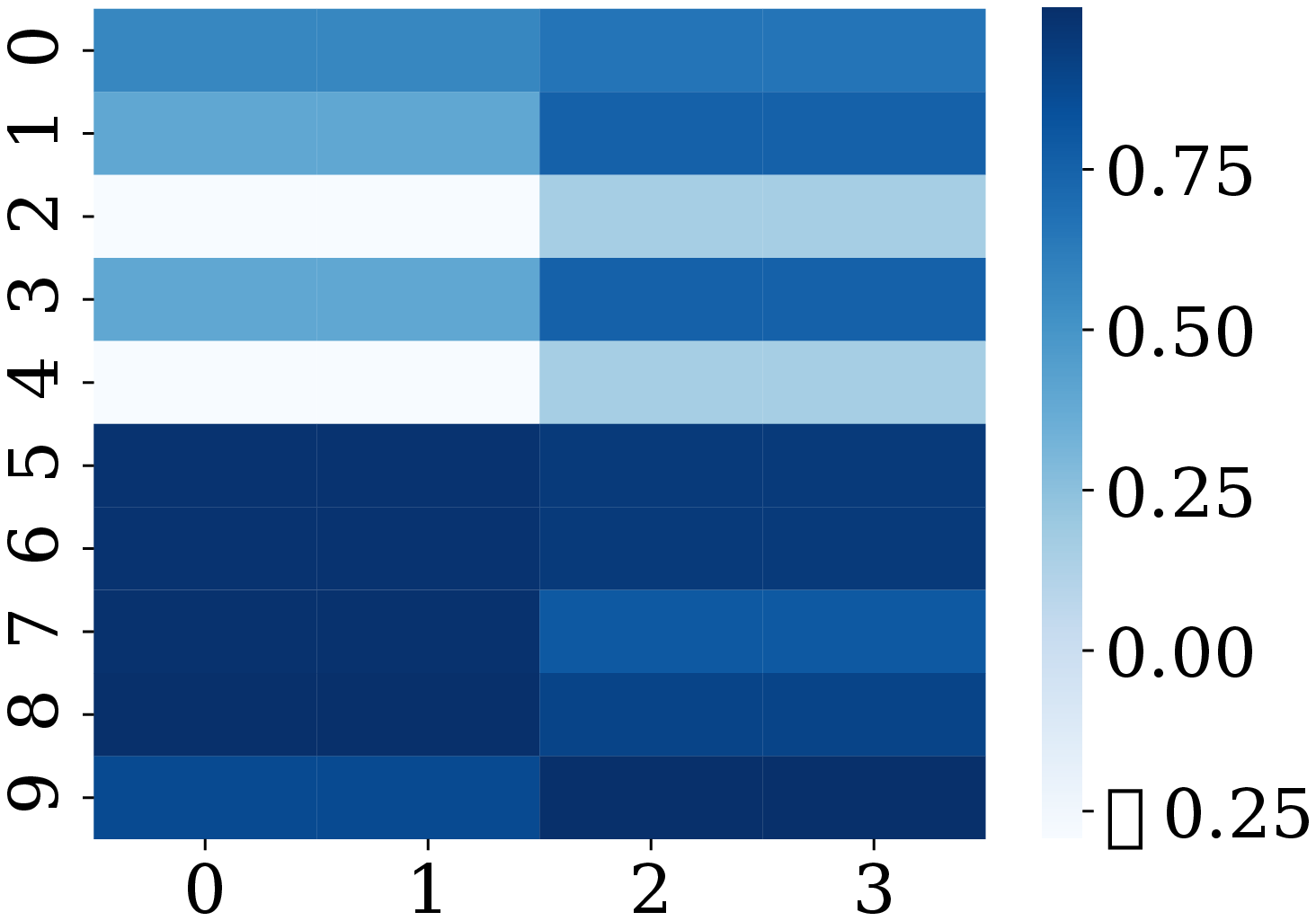}}}
    \vspace{-0.0in}
    \caption{A query case study on LINUX.}
    \label{fig:search_demo_extra_5}
\end{figure}

\begin{figure}
    \centering
    \subfloat[Query ranking result.]
    {{\includegraphics[scale=0.57]{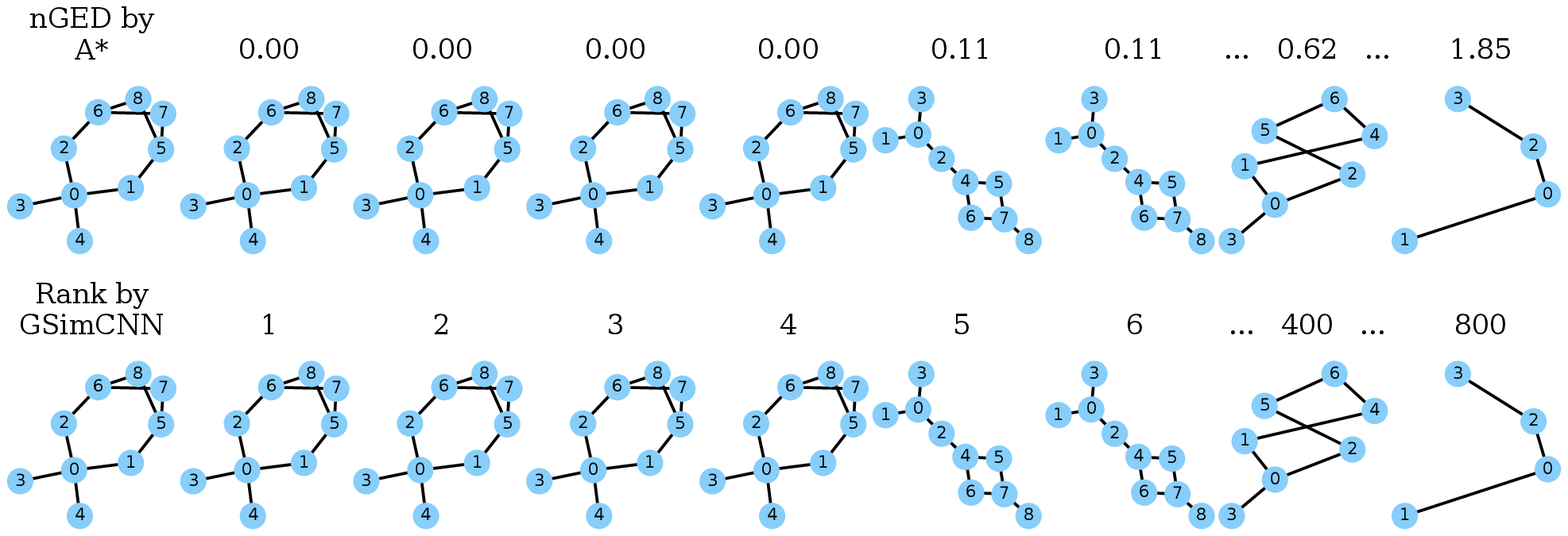}}}
    \vspace{-0.0in}
    \centering
    \subfloat[Similarity matrix between the query and the most similar graph ranked by both methods.]
    {{\includegraphics[scale=0.38]{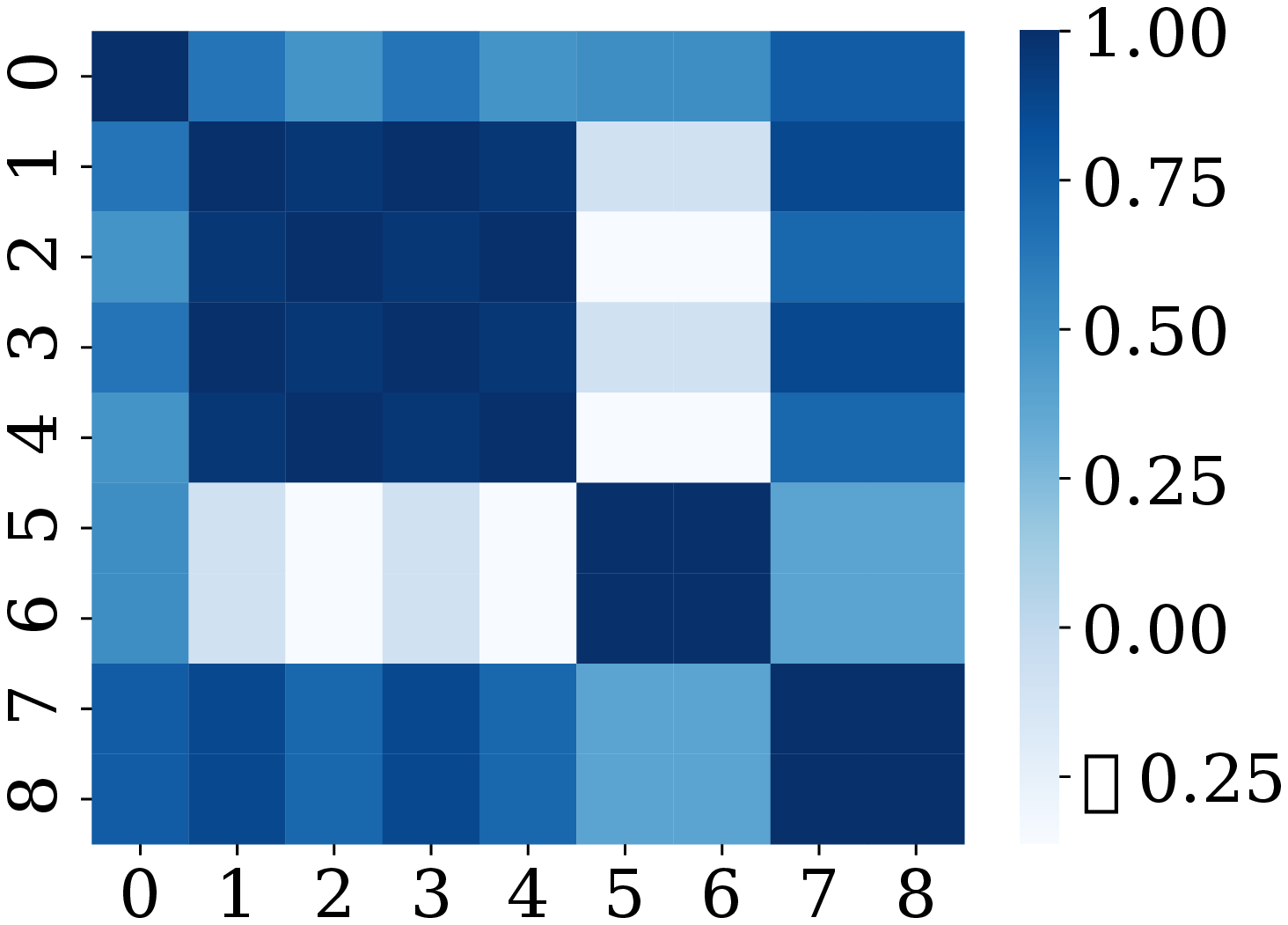}}}
    \vspace{-0.0in}
    \centering
    \subfloat[Similarity matrix between the query and the least similar graph ranked by both methods.]
    {{\includegraphics[scale=0.38]{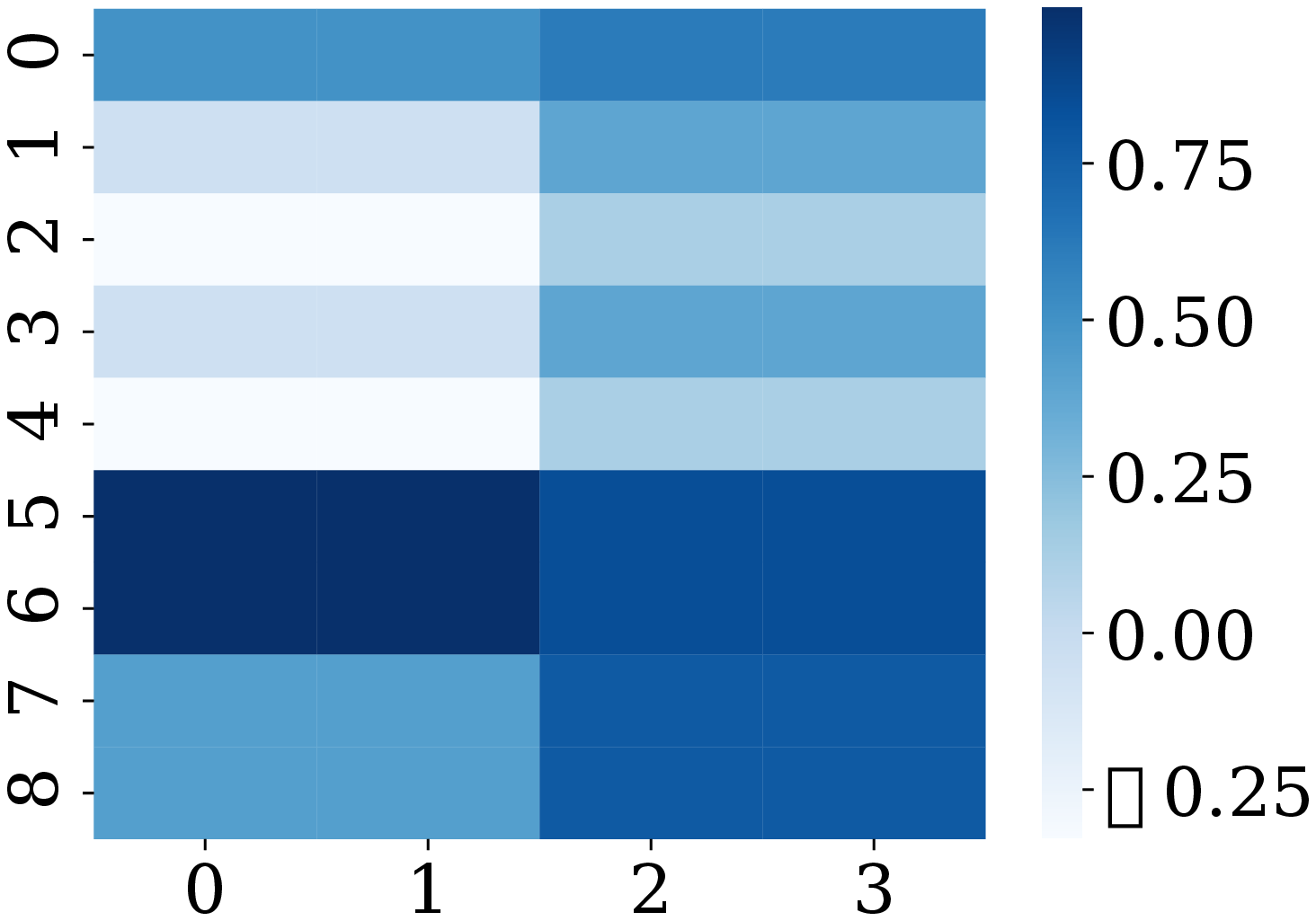}}}
    \vspace{-0.0in}
    \caption{A query case study on LINUX.}
    \label{fig:search_demo_extra_6}
\end{figure}

A few more visualizations are included in Fig.~\ref{fig:search_demo_extra_1}, \ref{fig:search_demo_extra_2}, \ref{fig:search_demo_extra_3}, \ref{fig:search_demo_extra_4}, \ref{fig:search_demo_extra_5}, and \ref{fig:search_demo_extra_6}. In each figure, two similarity matrices are visualized, the left showing the similarity matrix between the query graph and the most similar graph, the right showing the similarity matrix between the query and the least similar graph. 

\end{document}